\RequirePackage{fix-cm}
\documentclass[twocolumn]{svjour3}          
\smartqed  
\usepackage{tikz,xcolor,hyperref}
\usepackage{graphicx}
\usepackage[misc]{ifsym}
\usepackage{mathptmx}      
\usepackage{latexsym}
\usepackage{amsmath}
\usepackage{subfigure}
\usepackage{tabularx}
\usepackage{multirow}
\usepackage{booktabs}
\usepackage{float}
\usepackage{stfloats}
\usepackage{ragged2e}
\usepackage{bookmark}
\usepackage{color}
\usepackage[latin1]{inputenc}
\renewcommand{\raggedright}{\leftskip=0pt \rightskip=0pt plus 0cm}

\journalname{Neural Computing and Applications}
\begin{document}

\title{SRA-LSTM: Social Relationship Attention LSTM for Human Trajectory Prediction}
\author{Yusheng Peng\textsuperscript{1,2} \and Gaofeng Zhang\textsuperscript{3} \and Jun Shi\textsuperscript{3} \and Benzhu Xu\textsuperscript{3} \and Liping Zheng\textsuperscript{1,2,3}}
\institute{
\begin{itemize}
      \item[\Letter] {Liping Zheng} \\
            \email{zhenglp@hfut.edu.cn}
      \at
      \item[\textsuperscript{1}] School of Computer Science and Information Engineering, Hefei University of Technology, Hefei, China
      \item[\textsuperscript{2}] Anhui Province Key Laboratory of Industry Safety and Emergency Technology, Hefei University of Technology, Hefei, China
      \item[\textsuperscript{3}] School of Software, Hefei University of Technology, Hefei, China
    \end{itemize}
}
\date{Received: date / Accepted: date}
\maketitle

\begin{abstract}
Pedestrian trajectory prediction for surveillance video is one of the important research topics in the field of computer vision and a key technology of intelligent surveillance systems. Social relationship among pedestrians is a key factor influencing pedestrian walking patterns but was mostly ignored in the literature. Pedestrians with different social relationships play different roles in the motion decision of target pedestrian. Motivated by this idea, we propose a Social Relationship Attention LSTM (SRA-LSTM) model to predict future trajectories. We design a social relationship encoder to obtain the representation of their social relationship through the relative position between each pair of pedestrians. Afterwards, the social relationship feature and latent movements are adopted to acquire the social relationship attention of this pair of pedestrians. Social interaction modeling is achieved by utilizing social relationship attention to aggregate movement information from neighbor pedestrians. Experimental results on two public walking pedestrian video datasets (ETH and UCY), our model achieves superior performance compared with state-of-the-art methods. Contrast experiments with other attention methods also demonstrate the effectiveness of social relationship attention.
\keywords{trajectory prediction \and social interaction \and social relationship \and social relationship attention}
\end{abstract}

\section{Introduction}
\label{intro}
Pedestrian trajectory prediction is an important research topic in the field of computer vision, and has various applications, such as intelligent surveillance system, autonomous driving system and robot navigation system. Intelligent surveillance system plan an important role in preventing and investigating crimes, protecting public safety, and safeguarding national security. However, due to the intricate and subtle interactions between pedestrians, the pedestrian trajectory prediction for surveillance video is still challenging, which has attracted much attention from both academia and industry in recent years.

In summary, the general framework of intelligent surveillance consists of several stages including environment modeling \cite{1}, object classification \cite{2}, tracking \cite{3}, behaviour understanding and description \cite{4}, motion understanding and detection \cite{5}, and possibly personal identification \cite{6}. Our work is concerned with the stage of behavior and motion analysis, and more specifically, predicting the future trajectory of given trajectory data from the tracking stage. Before executing the trajectory prediction task, the position sequences of all pedestrians in the video are obtained through the tracking algorithm \cite{7}.

The trajectory of a pedestrian can be influenced by multiple factors such as scene topologies, pedestrian beliefs, and human-human interactions. Many scholars have worked to model the intricate and subtle interactions that often take place among the pedestrians. Although the hand-crafted energy functions adopted in earlier works \cite{8,9,10,11} attempted to build crowd interactions among pedestrians in crowded spaces, it is challenging to take a comprehensive consideration of various social behaviors. In recent years, Long-Short Term Memory networks(LSTM) \cite{12} are widely used for each pedestrian to capture latent motion patterns in the trajectory prediction task. In LSTM architecture based trajectory prediction methods, the pooling \cite{13,14}, attention \cite{15,16}, and graph \cite{17,18} mechanisms are widely used to model complex social interactions. The pooling mechanism utilizes social pooling on the occupancy maps to integrate the latent motions of pedestrians in a local neighborhood or the whole scene. The attention mechanism integrates latent motions by assigning different attention weights to neighbors. The graph structure mechanism constructs a graph based on observed trajectories of participants to explicitly model the paired interactions between participants. Compared with the other two mechanisms, the attention mechanism can model the contributions of different neighbor pedestrians to the target pedestrian to better predict the future trajectory. For this reason, some pooling based \cite{19,22} and graph based \cite{20} approaches are coupled with attention mechanism to achieve good prediction performance. The attention mentioned in existing work is always captured by latent motion patterns \cite{15,21} or spatial information(i.e. distance, relative position, bearing angle) \cite{16,22}. However, both types of attention methods simply capture attention from possible motion interactions.
\begin{figure}[pt!]
\centering
\includegraphics[scale=0.28]{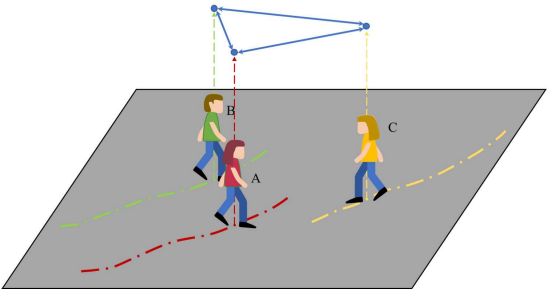}
\caption{Illustration of the social relationship attention in the crowd. A and B are friends with similar motion patterns of their trajectories. A and C are strangers who come from different source places and go to different destinations. The social relationships of A-B and A-C are different so that A pays different attention to B and C when predicting the future trajectory.}
\label{Fig:1}
\end{figure}

However, although the diverse aspects have been well researched, a factor was neglected in previous work. The inference of the pedestrian's future trajectory is not only affected by the motion interaction but also closely related to the social relationship between pedestrians. As shown in Fig.\ref{Fig:1}, A and B are friends, and the motion patterns of their trajectories are similar. Because of their accompanying behavior, they would pay higher attention to each other when predicting the future trajectory. As to A and C, who have a strange relationship, they keep less attention to each other unless there has possible motion interaction. We intend to model social interaction by social relationship attention that coupling the social relationship between pedestrians and possible motion interactions. Social relationship attention takes into account the influence of social relationships and motion interactions on pedestrian trajectory prediction.

To model this new insight, we present a novel LSTM architecture that recursively encodes social relationships and integrates information by social relationship attention, called Social Relationship Attention LSTM (SRA-LSTM). For each pedestrian, an LSTM is used to capture latent motion patterns. For each pair of pedestrians, an LSTM is applied to encode their relative positions and thus generate the feature representation for manifesting their social relationship. The social relationship attention is captured by the hidden state of motion LSTM and social relationship encoder, which used to integrate neighbor pedestrians' hidden states. The integrated hidden state is served as the input of motion LSTM, and the output hidden state predicts the position of the next key frame. Experiments on multiple human trajectory baseline, including two datasets in ETH \cite{9} and three datasets in UCY \cite{23}, show the superiority of our model. Our contributions can be summarized as follows:
\begin{enumerate}
\renewcommand{\labelenumi}{\theenumi.}
  \item We propose a novel interactive recurrent structure, namely SRA-LSTM, serving as a new pipeline for jointly predicting the future trajectories of pedestrians in the crowd.
  \item We adopt an LSTM as a social relationship encoder to model the temporal correlation of the relative position among pedestrians to obtain a representation of their social relationship. To the best of our knowledge, the modeling of social relationships by the temporal correlation of relative positions hasn't been considered separately yet.
  \item We propose a novel attention mechanism called social relationship attention, which takes into account the consideration of the social relationship between pedestrians and their motion patterns. This attention module is applied to the prediction task, which contributes to greater expressive power and higher performance.
\end{enumerate}

The rest of the paper is organized as follows. Section \ref{sec2} reviews the related works. The proposed SRA-LSTM model for human trajectory prediction is described in Section \ref{sec3}. Experiment results are evaluated and discussed in Section \ref{sec4}. We conclude this paper and point out the future works in Section \ref{sec5}.

\section{Related Works}\label{sec2}

The social interaction in trajectory prediction contains Human-human interaction and Human-Scene interaction. The former models the dynamic content of the scenes, i.e. how pedestrians interact with each other. The latter tends to learn scene-specific motion patterns \cite{24,25,26,27,28}. The focus of our work is the former: learning to model human-human interactions. We review existing work on this topic as well as relevant work in LSTMs based and attention approaches in trajectory prediction.

\subsection{LSTM models for trajectory prediction}\label{subsec2.1}
As a classic variant of RNNs, LSTM has achieved great success in various sequence tasks \cite{29,30,31}. Aliha \cite{13} proposed the Social-LSTM model that first introduced the LSTM network for pedestrian trajectory prediction. In Social-LSTM model, a separate LSTM network is used for each trajectory in a scene, and the LSTMs are connected through a social pooling layer. These LSTM framework models\cite{32,33} utilize LSTMs to capture motion features and connect the LSTMs through specific designed modules to model social interaction among pedestrian

Besides being the main framework of prediction models, the LSTM network is often used as the encoder and decoder in GAN framework models \cite{14,34,35} and encoder-decoder framework models \cite{36,37}. Among them, the LSTM encoder is applied to learn motion features from historical trajectories, and the decoder acts as a predictor to predict future trajectories. Not limited to this, Huang et al. \cite{38} adopt an extra LSTM to encode the temporal correlations of interactions,  Xue et al. \cite{39,40} utilize two LSTM layers to capture the location information as well as the velocity information, and Syed et al. \cite{26} utilize three LSTMs to encode person, social, and scene scale information. In our approach, we utilize one LSTM to capture motion features, and another LSTM is served as a social relationship encoder to acquire the representation of social relationships between each pair of pedestrians.

\subsection{Research on human-human Interactions}\label{subsec2.2}
Modeling social interactions is crucial for pedestrian trajectory prediction. These interactions can be defined by hand-crafted rules, such as social forces \cite{8} and stationary crowds¡¯ influences \cite{10,11}. However, it is difficult to capture the complex interactions in the scenario by these rules. Therefore, the majority of current research captures social interactions in a data-driven manner. Social-LSTM \cite{13} firstly proposes the social pooling layer on the occupancy maps to aggregate neighbor's latent motion dynamics in a local neighborhood. Social-GAN \cite{14} removes the regional restraint and directly aggregates the latent motions of neighbors in the whole scene.

Different from the pooling mechanism, the attention-based message-passing mechanism models \cite{39,40,41} aggregates information by paying different attention to neighbor pedestrians, which can better model social interaction. But beyond that, some recent models \cite{17,34,38} consider the pedestrians in a scene as nodes on the graph and leverage the Graph Convolution Networks (GCNs) or Graph Attention Networks (GATs) to aggregate information from neighbors. Nevertheless, we still apply the attention mechanism to model social interaction and propose a novel attention.

\subsection{Attention approaches for trajectory prediction}\label{subsec2.3}
Attention mechanisms have been proven to be significantly successful in various tasks \cite{42,43,44,45}. Fernando et al. \cite{15} first apply the attention mechanism to pedestrian trajectory prediction tasks, in which soft attention is acquired through hidden states and hard attention is calculated from distance. Analogously, the soft and hard attention in TPPO model \cite{22} are calculated on the basis of the cosine values of all pedestrians' bearing angles. However, Amirian et al. \cite{19}coupling multiple spatial features¡ªEuclidean distance, bearing angle, and closest distance¡ªto capture the attentional weights. Zhang et al. \cite{33} combines the hidden states of each pair of pedestrians and the embedding vector of their relative position to capture attentional weights.

Besides the above attention approaches, the temporal attention mechanism captures the relationships between a time step of the prediction phase and different time steps of the observed part of the input trajectory. Xue et al. \cite{40} apply two temporal attention mechanisms to the hidden states from the location and velocity LSTM layers. Zhao et al. \cite{37} apply a spatial-temporal attention mechanism to connect the decoder state and the temporal encoder state, which allows give an importance value for each time instant's trajectory state of the neighboring humans and the target human. In our approach, we do not directly capture spatio-temporal attention, but acquire spatio-temporal features through a specific encoder, and then couple these features with hidden states to capture attention weights.

Our work in this paper differs from the above works in two main aspects: firstly, the above works model social interactions by the spatial relationship among pedestrians and latent motions while we introduce social relationships to model social interactions. Secondly, we did not adopt the manual label 0/1 in \cite{18} to represent the social relationship among pedestrians, but utilize the LSTM model to actively learn deeper representation of social relationships from the relative movement information among pedestrians. 

\begin{figure*}[ht]
\centering
\includegraphics[scale=0.36]{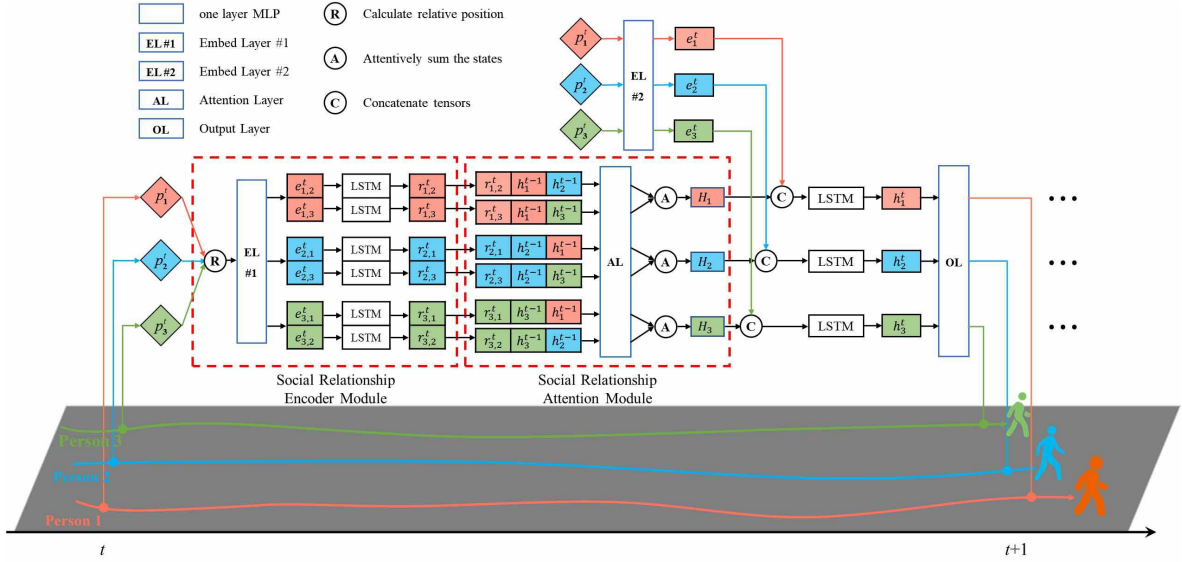}
\caption{Illustration of the overall approach. At each time-step, the pedestrians positions are used to calculate relative positions of each other, and the relative positions are processed through embed layer and LSTM to encode the social relationships of each pair of pedestrians. The social relationship attention module models the social interactions by attentively integrating the hidden states of neighbors. Then the social interaction tensor and the embedding vector of each pedestrian's position are treated as inputs of LSTM to output the current hidden states and infer the positions of next time-step.}
\label{Fig:2}
\end{figure*}

\section{Proposed Method}\label{sec3}

The overview of the SRA-LSTM framework is illustrated in Fig.\ref{Fig:2}. We treat one kind of LSTM as a social relationship encoder to model the temporal correlation of the relative position between pedestrians to obtain a representation of their social relationships. Then, the social interactions are modeled by a social relationship attention module that captures social relationship attention and attentively integrates hidden states which represent the latent motion patterns. The embedding of position and output social tensor are treated as an input of motion LSTM for each person to capture current latent motions and infer the position of the next time-step.

\subsection{Problem Formulation}\label{subsec3.1}

In this paper, we address the problem of pedestrian trajectory prediction in the crowd scenes for surveillance video. For better modeling the social interactions among pedestrians, we focus on two-dimensional coordinations of pedestrians in the world coordinate system at specific key frames. For each sample, we assume there are \emph{N} pedestrians involved in the scene of surveillance video. Given certain observed positions $\{p_i^t|(x_i^t, y_i^t), t=1,2,...,T_{obs}\}$ of pedestrians \emph{i} of $T_{obs}$ key frames, our goal is predicting the positions $\{p_i^{t'}|(\widehat{x}_i^{t'}, \widehat{y}_i^{t'}),{t'}=T_{obs}+1,T_{obs}+2,...,T_{pred}\}$ of future $T_{pred}$ key frames.

\subsection{Social Relationship Encoding}\label{subsec3.2}
In crowd scenes, pedestrians with different social relationships exhibit different social behaviors. Pedestrians with intimate relationships (such as lovers, friends) tend to maintain a stable intimate distance while walking. Pedestrians tend to change their walking routes to avoid the approaching of strange pedestrians. Therefore, the social relationship between pedestrians is a factor that cannot be ignored in pedestrian trajectory prediction modeling. However, the social relationship between a pair of pedestrians is unknown in the trajectory prediction task. Our primary task is acquiring knowledge that can represent the social relationship between pedestrians.

\begin{figure*}[ht]
\centering
\includegraphics[scale=0.38]{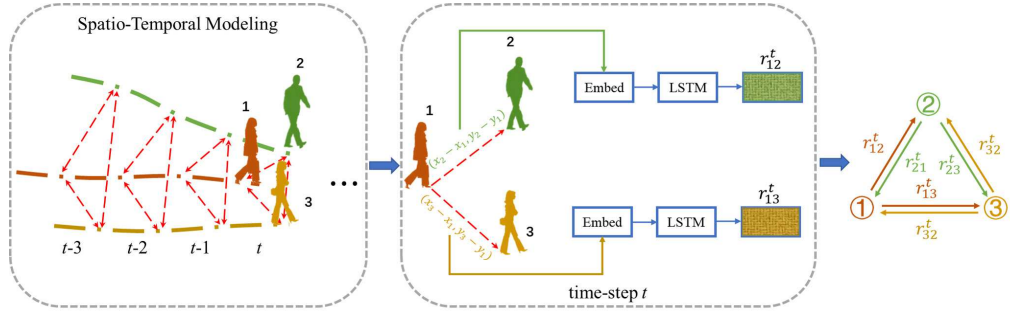}
\caption{We model the spatio-temporal relationship of pedestrians' positions to capture the representation of social relationship between each pair of pedestrians. For each time-step, the relative position between couple of pedestrians is processed through embed layer and LSTM to output the representation tensor of social relationships.}
\label{Fig:3}
\end{figure*}

Motivated by the idea above, Sun et al. \cite{18} utilize group-based social interaction model to explore relationships among pedestrians. Sun invited experts who have sociological background to judge the relationship of two pedestrians and utilize labels to annotate whether pedestrians belong to the same group. In contrast, we intend to capture the representation of social relationship by a spatio-temporal model. Therefore, we try to obtain a representation of the social relationship by modeling the temporal correlation of the relative positions between pedestrians. In our method, for each pair of pedestrians, an LSTM is used as an encoder to encode the relative position of the pedestrians recurrently to capture the representation of their social relationship. As illustrated in Fig.\ref{Fig:3}, for the target pedestrian i, the relative positions of the neighbor pedestrians are calculated and served as the current input of the LSTM encoder. The current hidden state of the LSTM encoder represents the social relationship between the pair of pedestrians. We term this LSTM as R-LSTM (LSTM for social relationship encoding):
\begin{align}
  & e_{ij}^t={\phi}_r(x_j^t-x_i^t, y_j^t-y_i^t;W_{re}) \label{eq:1} \\
  & r_{ij}^t=\textrm{R-LSTM}(r_{ij}^{t-1},e_{ij}^t;W_r) \label{eq:2}
\end{align}
where $e_{ij}^t$ is embedding vector of the relative spatial location. $W_{re}$ denotes the parameter for the embedding function ${\phi}_r$. $r_{ij}^t$ is the hidden state of the R-LSTM among pedestrian \emph{i} and \emph{j} at the time-step \emph{t}, $W_r$ is the R-LSTM weight and is shared among all the sequences.

In real life interpersonal communication, the intimate relationship between people is not equal because of the diversity and the difference in interpersonal emotions. As illustrated by the right part of Fig.\ref{Fig:3}, the understanding of social relationships between pedestrians is not equal, e.g. $r_{12}^t \neq r_{21}^t$. $r_{12}^t$ represents the pedestrian 1's understanding of the social relationship between pedestrians 1 and 2. Similarly, $r_{12}^t$ is the understanding of social relationship from pedestrian 2. The representation $r_{ij}^t$ of the social relationship between pedestrians is continuously learning as time steps increases.

\subsection{Social Relationship Attention Module}\label{subsec3.3}

\begin{figure}[btp!]
\subfigure[]{
\centering
\includegraphics[width=3.8cm]{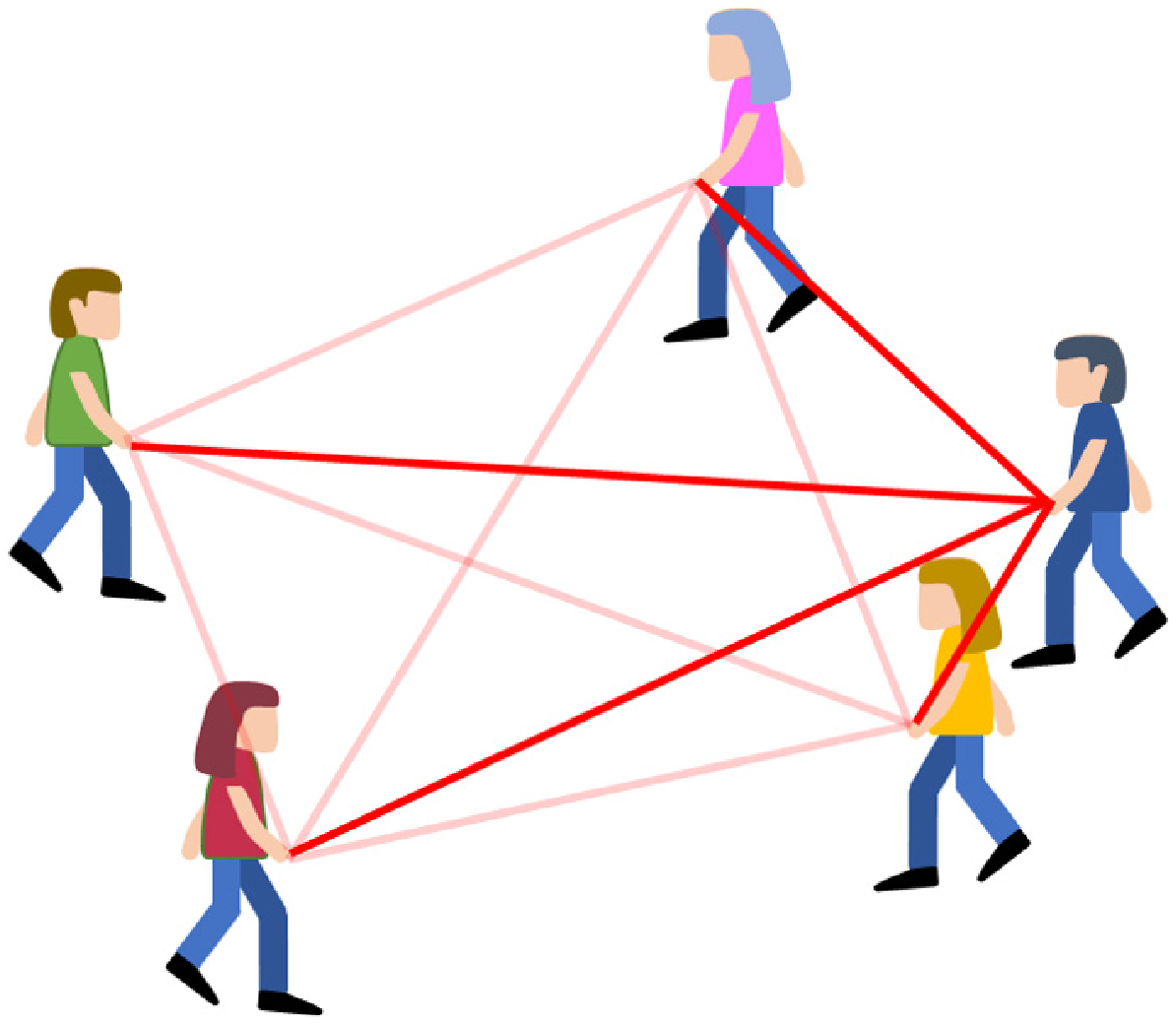}
\label{Fig:4a}}
\subfigure[]{
\centering
\includegraphics[width=3.8cm]{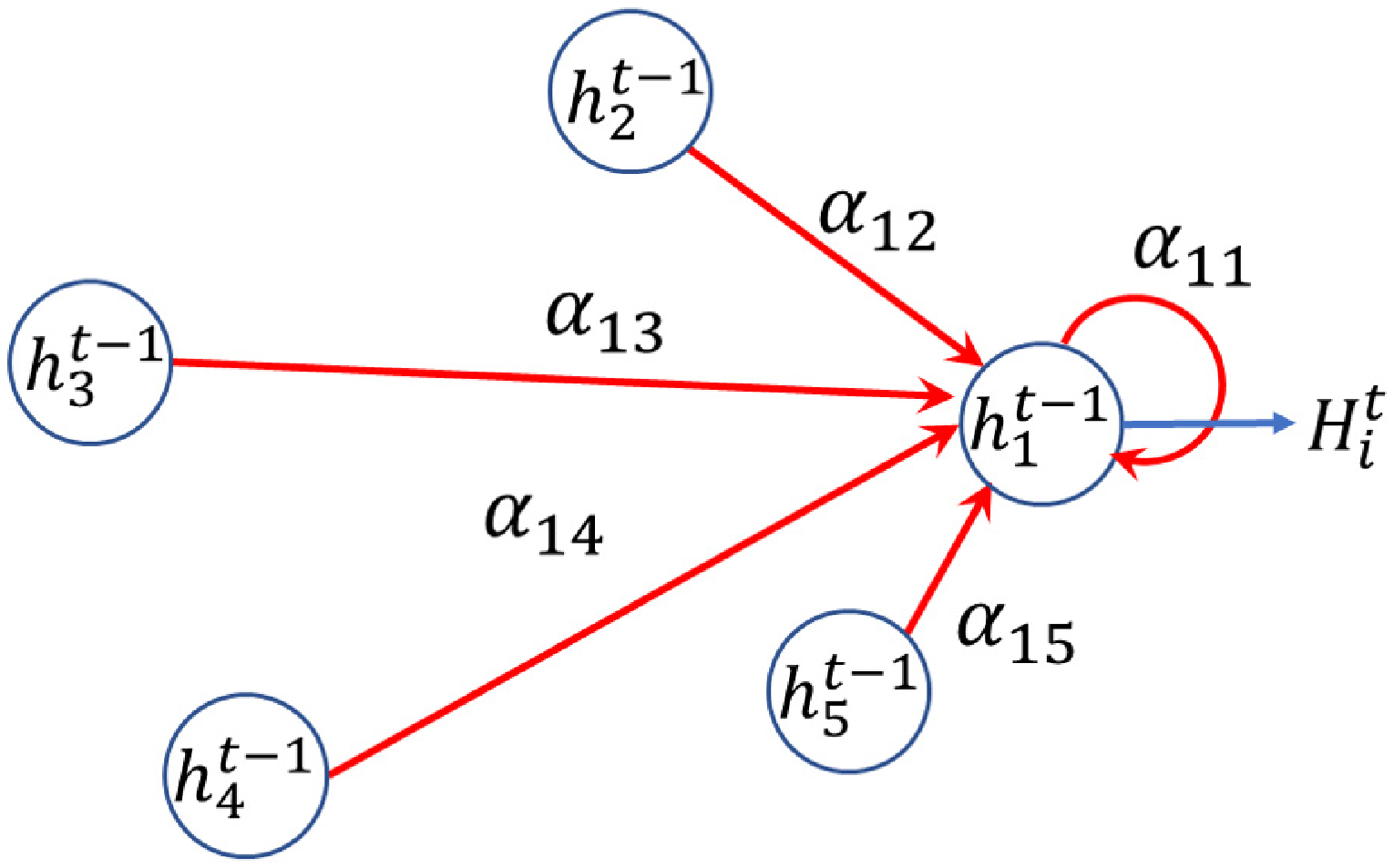}
\label{Fig:4b}}
\centering
\caption{(a) The red lines between pedestrians represent the exist of human-human interactions. (b) The target pedestrian attentively aggregate the features of different neighbors with different importance.}
\label{Fig:4}
\end{figure}

Vanilla LSTM used for per person does not capture human-human interactions. To model the social interaction behavior between pedestrians, pooling mechanism \cite{13,14,32} is used to aggregate hidden states between pedestrians on occupancy map. Recently, the trajectory prediction models \cite{15,33,37} that adopt attention mechanisms to model social interaction have achieved good performance. This is because the attention mechanism can pay different attention to aggregate the shared information according to the importance of neighbor pedestrians. Therefore, we adopt the attention mechanism to model social interactions. SR-LSTM \cite{33} utilizes relative position and hidden states to capture attention in state refinement module to model social interactions. ST-Attention \cite{37} adopts soft attention \cite{15} to acquire social interaction features in encoder, and spatial attention is adopted to capture the different influences of historical location sequences on future trajectories in the decoder. Differ from them, we utilize social relationship representation to capture attention which has similar temporal-spatial property. As shown in Fig.\ref{Fig:4}(a), Red lines between pedestrians represent the existence of human-human interactions.

The location and movement intention of the neighbor pedestrian are the key influence to the decision of the target pedestrian's movement. Furthermore, pedestrians with different social relationships always affect the motion decision of the target pedestrian. Motivated by the idea, we introduce a novel attention called social relationship attention which is captured by coupling pedestrians' latent motion patterns and social relationships of each pair of pedestrians. The social relationship attention is calculated by:
\begin{equation}\label{eq:3}
  \alpha_{ij}^t = \frac{\exp{W^{at}[r_{ij}^t;h_i^{t-1};h_j^{t-1}]}}{\sum_{k\in N(i)}{\exp{W^{at}[r_{ik}^t;h_i^{t-1};h_k^{t-1}]}}}
\end{equation}
where the $r_{ij}^t$ is the hidden state of the R-LSTM, which represents the social relationship between pedestrians \emph{i} and \emph{j}. The $h_i^{t-1}$ and $h_i^{t-1}$ are hidden states of pedestrian \emph{i} and \emph{j} at time-step $t-1$, which represent the latent motion patterns of \emph{i} and \emph{j} respectively. $W^{at}$ are a weight matrix.

After capturing the social relationship attention, we aggregate the latent motion patterns by paying attention to neighbor pedestrians. As illustrated in Fig.\ref{Fig:4}(b), we attentively aggregate the hidden states of each actor to obtain social interaction context. The social interaction context vector for pedestrian \emph{i} at time-step \emph{t} is given by:
\begin{equation}\label{eq:4}
  H_i^t = \sum_{j\in N(i)}{{\alpha_{ij}^t}h_j^{t-1}}
\end{equation}
where $h_j^{t-1}$ is the hidden state encoded by history positions of pedestrian \emph{j}. $N(i)$ is the set of neighbors of pedestrian \emph{i}, and $\alpha_{ij}^t$ is the attention weight of the pedestrian pair $(i, j)$.

\subsection{SRA-LSTM Prediction Model}\label{subsec3.4}

LSTM has been proven to successfully capture the historical motion state of a single pedestrian \cite{13,32,33}. By following these works, we also employ an LSTM denoted as motion LSTM to capture the latent motion pattern for each pedestrian. In our implementation, we use the normalized absolute(Nabs) position \cite{33} which shifts the origin to the latest observed time slot:
\begin{equation}\label{eq:5}
  \begin{split}
  \Delta x_i^t&=x_i^t-x_i^{T_{obs}} \\
  \Delta y_i^t&=y_i^t-y_i^{T_{obs}}
  \end{split}
\end{equation}

The hidden state of motion LSTM of each pedestrian captured by his/her past Nabs positions, which represents the current latent motion pattern and infer the position of the next time-step. The latent motion patterns of pedestrians are independent of each other, so that the predicted positions may collide. Therefore, we model social interactions by social relationship attention module that integrate information of neighbor pedestrians for capturing a social context tensor. We treat the social context tensor and the embedding vector $e_i^t$ of position as inputs of the LSTM cell. This introduces the following recurrence:
\begin{align}
  & e_i^t=\phi(\Delta x_i^t, \Delta y_i^t;W_e) \label{eq:6} \\
  & h_i^t=\textrm{LSTM}(h_i^{t-1},H_i^t,e_i^t;W_l) \label{eq:7}
\end{align}
where $\phi(\cdot)$ is an embedding function with ReLU nonlinearity, $W_e$ is the embedding weights. The LSTM weight is denoted by $W_l$. These parameters are shared among all the pedestrians in the whole scene. $H_i^t$ is the social context tensor which is the output of the social relationship attention module. It is the social interaction feature which is integrate the hidden states of the interested person and neighbors.

The hidden-state $h_i^t$ at time \emph{t} is used to predict the Nab position $(\Delta \widehat{x}_i^{t+1}, \Delta \widehat{y}_i^{t+1})$
at the next time-step \emph{t+1}:
\begin{equation}\label{eq:8}
  [\Delta \widehat{x}_i^{t+1}, \Delta \widehat{y}_i^{t+1}]^T=W_ph_i^t
\end{equation}
where $W_p$ is a weight matrix. From time $T_{obs}+1$ to $T_{pred}$, we transform the predicted Nabs positions to absolute positions, which utilized to calculate relative positions to encode social relationship.

\subsection{Implementation Details}\label{subsec3.5}
The parameters of the SRA-LSTM model are directly learned by minimizing the L2 loss between the predicted position and ground truth. All LSTMs in our implementation only have one layer. The dimension of hidden states of all LSTM cells is set to 64. The dimension of embed vector $e_{ij}^t$ in Eq.\ref{eq:1} and $e_i^t$ in Eq.\ref{eq:6} are set to 32. A sliding time window with a length of 20 and a stride size of 1 is adopted to get the training samples. All trajectory segments in the same time window are regarded as a mini-batch, as they are processed in parallel. Adam optimizer is adopted to train models in 300 epochs, with an initial learning rate of 0.001.

\begin{table*}[!ht]
\caption{Quantitative results of all the baselines and the proposed method (SAM:Samples, SCE:Scene, ATT:Attention, SRS:Social Relationships)}\label{tab1}
    {\begin{tabular*}{\textwidth}{@{\extracolsep{\fill}}c|cccc|cccccc}
    \toprule
    \multirow{2}[4]{*}{Method} & \multicolumn{4}{c|}{Notes} & \multicolumn{6}{c}{Performance(ADE/FDE)} \\
\cmidrule{2-11}          & SAM & SCE & ATT & SRS & ETH-univ & ETH-hotel & UCY-zara01 & UCY-zara02 & UCY-univ & AVG \\
    \midrule
    \midrule
    S-LSTM \cite{13} &       &       &       &       & 1.09 / 2.35 & 0.79 / 1.73 & 0.47 / 1.00 & 0.56 / 1.17 & 0.67 / 1.40 & 0.72 / 1.54 \\
    SGAN \cite{14} & $\surd$ &       &       &       & 0.87 / 1.62 & 0.67 / 1.37 & 0.35 / 0.68 & 0.42 / 0.78 & 0.76 / 1.52 & 0.61 / 1.21 \\
    Sophie \cite{35} & $\surd$ & $\surd$ & $\surd$ &       & 0.70 / 1.43 & 0.76 / 1.67 & \textbf{0.30 / 0.62} & 0.38 / 0.75 & 0.54 / 1.24 & 0.54 / 1.15 \\
    S-BiGAT \cite{34} & $\surd$ & $\surd$ & $\surd$ &       & 0.69 / 1.29 & 0.49 / 1.01 & \textbf{0.30 / 0.62} & 0.36 / 0.75 & 0.55 / 1.32 & 0.48 / 1.00 \\
    SR-LSTM \cite{33} &       &       & $\surd$ &       & 0.63 / 1.25 & 0.37 / 0.74 & 0.41 / 0.90 & 0.32 / 0.70 & \textbf{0.51 / 1.10} & \textbf{0.45} / 0.94 \\
    ST-Attenion \cite{37}  &      &     & $\surd$ &       & 0.85 / 1.85 & 0.32 / 0.66 & 0.42 / 0.91 & 0.34 / 0.73 & 0.63 / 1.33 & 0.51 / 1.10 \\
    RSBG \cite{18} &       &       &       & $\surd$ & 0.80 / 1.53 & 0.33 / 0.64 & 0.40 / 0.86 & \textbf{0.30 / 0.65} & 0.59 / 1.25 & 0.48 / 0.99 \\
    \midrule
    \midrule
    Ours  &       &       & $\surd$ & $\surd$ & \textbf{0.59 / 1.16} & \textbf{0.29 / 0.56 } & 0.37 / 0.82 & 0.43 / 0.93 & 0.55 / 1.19 & \textbf{0.45 / 0.93} \\
    \bottomrule
    \end{tabular*}}{}
\end{table*}%

\section{Experiments}\label{sec4}

In this section, we evaluate our method on two public walking pedestrian video datasets: ETH and UCY. These two datasets contain 5 crowd scenes, including ETH-univ, ETH-hotel, UCY-zara01, UCY-zara02, and UCY-univ. There are 1536 pedestrians and thousands of real-world pedestrian trajectories. All the trajectories are converted to the world coordinate system and then interpolated to obtain values at every 0.4 seconds.

\textbf{Evaluation Metrics}. Similar to prior works \cite{13,14}, the proposed method is evaluated with two types of metrics as follow:
\begin{enumerate}
\renewcommand{\labelenumi}{\theenumi.}
  \item \emph{Average Displacement error(ADE)}: the mean square error(MSE) between the ground-truth trajectory and predicted trajectory over all predicted time steps.
  \item \emph{Final Displacement error(FDE)}: the mean square error(MSE) between the ground-truth trajectory and predicted trajectory at the last predicted time steps.
\end{enumerate}

\textbf{Baseline}. We do not use traditional methods(such as linear model, social force model, etc.) based on handcrafted features as baselines, but only compared with the deep learning models. We compare the proposed model with the following recent works:
\begin{enumerate}
  \item \emph{S-LSTM} \cite{13}: An improved trajectory prediction method that combines LSTM with a social pooling layer, which can aggregate hidden states of the neighbor pedestrians. Future trajectories are predicted by decoding the concatenation of coordinates embedding and social pooling vector.
  \item \emph{SGAN} \cite{14}: An improved version of S-LSTM that the social pooling is displaced with a new pooling mechanism which can learn a "global" pooling vector. A variety loss function is proposed to encourage the GAN to spread its distribution and generate multiple socially acceptable trajectories.
  \item \emph{SoPhie} \cite{35}: An attentive GAN model for trajectory prediction by using social and physical attention modules. The trajectory prediction performance is improved by highlighting the key information with attention operations.
  \item \emph{S-BiGAT} \cite{34}: An improved version of SGAN by using the bicycle structure to train the generator. A graph attention network is used to model social interactions for better prediction performance. Self-attention is used on images to consider the physical feature of a scene.
  \item \emph{SR-LSTM} \cite{33}: An improved version of S-LSTM by proposing a data-driven state refinement module. The refinement module can jointly and iteratively refines the current states of all participants in the crowd on the basis of their neighbors' intentions through a message passing mechanism.
  \item \emph{ST-Attention} \cite{37}: A novel spatial-temporal attention (ST-Attention) model, which studies spatial and temporal affinities jointly. Among them, a novel temporal attention mechanism is used to extract temporal affinity, and the soft attention mechanism \cite{15} is used to explore spatial affinity.
  \item \emph{RSBG} \cite{18}: A novel structure called Recursive Social Behavior Graph, which is supervised by group-based annotations. The social interactions are modeled by GCNs that adequately integrate information from nodes and edges in RSBG.
\end{enumerate}

\textbf{Evaluation Methodology}. We use the leave-one-out approach similar to that from S-LSTM \cite{13}. Specifically, we train models on four datasets and test on the remaining dataset. We take the coordinates of 8 key frames (3.2s) of the pedestrian as the observed trajectory, and predict the trajectory of the next 12 key frames (4.8s). For each mini-batch, random rotation is employed for data augmentation.

\subsection{Quantitative Evaluations}\label{subsec4.1}
\subsubsection{Evaluations on ADE/FDE Metrics}
We compare our method to the various baselines in Table \ref{tab1} and report the average displacement error (ADE) and final displacement error (FDE). We compare the differences among the methods in the notes section. SAM denotes that the model was trained using variety loss and predicts 20 samples for each observed trajectory during test time, ATT denotes using attention mechanism, SCE denotes using scene information, SRS denotes that the model take advantage of social relationship. Among these baselines, S-LSTM, SR-LSTM, ST-Attention, and RSBG are deterministic models. For these models, only one socially acceptable path is predicted of each person according to observed trajectory. The proposed SRA-LSTM is also a deterministic model. The SGAN, Sophie, S-BiGAT are stochastic models that can generate multiple socially acceptable future paths. The ADE and FDE of these methods are evaluated using 20 samples. The S-LSTM and SGAN adopt pooling mechanisms to model social interactions. As classic baselines, the results of these two models are not as well as other recent methods.

Different from S-LSTM and SGAN models, Sophie, S-BiGAT, SR-LSTM, and ST-Attention adopt attention mechanism to model social interactions. Besides this, Sophie and S-BiGAT take advantage of scene information to model the interactions between human and scene. While the scene information was not used in our model, its prediction performance was best on the ETH-univ and ETH-hotel datasets. Furthermore, the average ADE(FDE) of our model on the five datasets was reduced by 16.67\%(19.13\%) and 6.25\%(7\%), respectively, compared with the two models. SR-LSTM and ST-Attention only utilize attention mechanism to model human-human interactions. The proposed model performs better than SR-LSTM and ST-Attention on the ETH-univ, ETH-hotel and UCY-zara01 datasets. Besides, the average ADE and FDE of the five datasets is also better.

It is worth noting that only RSBG and our model regard social relationship as an impact factor of social interaction. RSBG introduces social related annotations and uses 0/1 to represent whether two pedestrians are in the same group. In contrast, our model acquire the representation of social relationship of each pair of pedestrians through a social relationship encoder from their relative positions. Due to the existence of stationary crowd groups in UCY-zara02 video, the RSBG model has the best performance on this dataset. But our model has better performance on the remaining datasets.

\subsubsection{Inference Speed and Model Size}
Table \ref{tab2} lists out the speed comparisons between our model and publicly available models which we could bench-mark against. The size of SRA-LSTM is 67.1K parameters. SGAN \cite{14} has the smallest model size with 46.4k parameters, which is about seven tenth of the number of parameters in SRA-LSTM. The size of SR-LSTM is 64.9K parameters which is very close to the number of parameters in SRA-LSTM. In terms of inference speed, SR-LSTM was previously the fastest method with an inference time of 0.0049 seconds per inference step. The inference time of our model is 0.0045 seconds per inference step which is about 1.09x faster than SR-LSTM. 

\begin{table}[!t]
	\centering
	\caption{Parameters size and inference time of different models compared to ours. The lower the better. Models were bench-marked using Nvidia GTX2080Ti GPU. The inference time is the average of several single inference steps. We notice that SRA-LSTM has the least inference time compared to others despite it has the most parameters. The text in blue show how many times our model is faster than others.}\label{tab2}
    {\begin{tabular*}{20pc}{@{\extracolsep{\fill}}l||ll}
		\toprule
		& Parameters count & Inference time \\
		\midrule
		SGAN \cite{14}  & \textbf{46.4K} (0.69x) & 0.0057 \textcolor{blue}{(1.27x)} \\
		\midrule
		SR-LSTM \cite{33} & 64.9K (0.97x) & 0.0049 \textcolor{blue}{(1.09x)} \\
		\midrule
		\midrule
		SRA-LSTM & 67.1K & \textbf{0.0045} \\
		\bottomrule
	\end{tabular*}}{}
\end{table}%

\subsection{Ablation Study}\label{subsec4.2}
In order to further illustrate the effectiveness of social relationship attention and the rationality of the key parameter settings of our model, we conduct ablation experiments. Our ablation experiments are also evaluated on the above five subsets, in terms of ADE and FDE. For each ablation experiment, we strictly control all the hyper-parameters to be consistent.
\subsubsection{Ablation Study of Different Attention Mechanisms}\label{subsec4.2.1}
In pedestrian trajectory prediction, pedestrian's motion decisions are affected by the scene layout and surrounding pedestrians' movement. Pedestrians with different distances, different social relationships, and different motion will have different influence on the target pedestrian's decision-making. Therefore, it is necessary to use attention mechanism to model social interaction. In our approach, we present a novel attention called "Social Relationship Attention(SRA)" which is acquired by coupling the latent motions of pedestrians and social relationship among them. We verify the effectiveness of social relationship attention through ablation study of different attention Mechanisms. Table \ref{tab3} shows the comparison of different attention methods. Column 2 in the table shows the trajectory prediction results with no attention. The soft attention \cite{15} denoted as SA is captured by the latent motions of the target pedestrian and his/her neighbor. We also adopt the SA to model social interactions in our comparison. Beside this, RA represents the attention in \cite{33} which is acquired from relative position and hidden states. According to the statistical results, the trajectory prediction based on SRA achieved the lowest ADE and FDE on ETH-univ, ETH-hotel, UCY-zara01, and UCY-univ datasets. As for the average ADE and FDE on the five datasets, the average ADE(FDE) of SRA is reduced by 13.46\%(23.94\%) and 11.76\%(14.68\%) compared to SA and RA. And compared with the model with no attention, the average ADE(FDE) of SRA-LSTM is reduced by 11.76\%(15.45\%).

\begin{table}[!t]
	\caption{Quantitative results of different types of attention methods in social attention module}\label{tab3}
	{\begin{tabular*}{20pc}{@{\extracolsep{\fill}}ccccc}
			\toprule
			\multirow{2}[4]{*}{Dataset} & \multicolumn{4}{c}{Attention type} \\
			\cmidrule{2-5}          & Non    & SA    & RA    & SRA \\
			\midrule
			\midrule
			ETH-univ & 0.69 / 1.38 & 0.64 / 1.27 & 0.67 / 1.35 & \textbf{0.59 / 1.16} \\
			ETH-hotel & 0.45 / 0.99 & 0.45 / 0.96 & 0.44 / 0.91 & \textbf{0.29 / 0.56} \\
			UCY-zara01 & 0.48 / 1.05 & 0.45 / 0.98 & 0.44 / 0.98 & \textbf{0.37 / 0.82} \\
			UCY-zara02 & \textbf{0.36 / 0.81} & 0.41 / 0.93 & 0.41 / 0.96 & 0.43 / 0.93 \\
			UCY-univ & 0.59 / 1.28 & 0.68 / 1.60 & 0.58 / 1.26 & \textbf{0.55 / 0.82} \\
			\midrule
			\midrule
			AVG   & 0.51 / 1.10 & 0.52 / 1.14 & 0.51 / 1.09 & \textbf{0.45 / 0.93} \\
			\bottomrule
	\end{tabular*}}{}
\end{table}%

\begin{table}[!t]
	\centering
	\caption{Quantitative results of SRA-LSTM model with different parameters. The value of dim1 is the dimention of all embedding vectors. The value of dim2 is the dimention of hidden states of all LSTM cells.}\label{tab4}
	\begin{tabular}{cc|cccc}
		\toprule
		\multirow{3}[4]{*}{Datasets} & \multicolumn{4}{c}{Key parameter settings} &  \\
		\cmidrule{2-6}          & dim1  & 32    & 32    & 64    & 64 \\
		& dim2  & 32    & 64    & 64    & 128 \\
		\midrule
		\multicolumn{2}{c|}{ETH-univ} & 0.64 /1.25 & \textbf{0.59}/1.16 & \textbf{0.59/1.14} & 0.63/1.23 \\
		\multicolumn{2}{c|}{ETH-hotel} & 0.32/0.61 & \textbf{0.29/0.56} & 0.31/0.62 & 0.31/0.66 \\
		\multicolumn{2}{c|}{UCY-zara1} & 0.44/0.96 & \textbf{0.43}/0.93 & \textbf{0.43}/0.94 & \textbf{0.43/0.92} \\
		\multicolumn{2}{c|}{UCY-zara2} & 0.37/0.82 & 0.37/0.82 & \textbf{0.36/0.79} & 0.38/0.88 \\
		\multicolumn{2}{c|}{UCY-univ} & 0.59/1.27 & \textbf{0.55/1.19} & 0.56/1.23 & 0.57/1.26 \\
		\midrule
		\midrule
		\multicolumn{2}{c|}{AVG} & 0.47/0.98 & \textbf{0.45/0.93} & \textbf{0.45}/0.94 & 0.46/0.99 \\
		\bottomrule
	\end{tabular}%
\end{table}%

\subsubsection{Ablation Study of Key Parameter Settings}\label{subsec4.2.2}
Table \ref{tab4} shows the comparison of key parameter settings of our SRA-LSTM model. The dim1 in table represent the dimension of embed vector $e_{ij}^t$ in Eq.\ref{eq:1} and $e_i^t$ in Eq.\ref{eq:6}. The dim2 represent the dimension of hidden states of all LSTM cells. As shown in Table \ref{tab4}, we list four groups parameter settings of dim1 and dim2 for ablation study. When dim1 and dim2 set to 32 and 64, the model achieves an average performance of ADE/FDE = 0.45/0.93 on the five subsets, which perform better than other parameter settings. The average ADE/FDE of the four groups of parameter settings is 0.4575/0.9600, and the standard deviation of ADEs/FDEs is 0.008/0.025. So it is clear that the change of dimensions has a small impact on the performance of the model. The main reason for this situation may be that the SRA-LSTM model is stable and robust.

\subsection{Qualitative Evaluations}\label{subsec4.3}

As mentioned before, the quantitative results show that SRA-LSTM outperforms state-of-art models in terms of ADE/FDE metrics. We now qualitatively analyze how the SRA-LSTM model successfully captures the social interactions. We compare the performances of SGAN, SR-LSTM, and our model in four common social scenarios. Besides, we provide comparisons between our model with SR-LSTM in two different social scenarios of group. Finally, we also analyzed the failure cases in two different scenarios.
\setcounter{subfigure}{-1}
\begin{figure*}[!t]
	\subfigcapskip=5pt
	\centering
	\subfigure
	{
		\centering
		\includegraphics[scale=0.4]{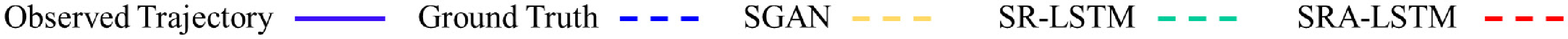}
	}
	\\
	\centering
	\subfigure[Parallel Walking]
	{
		\centering
		\begin{minipage}[c]{0.23\linewidth}
			\centerline{ETH-univ\_\#4277}
			\vspace{0.05cm}
			\includegraphics[scale=0.34]{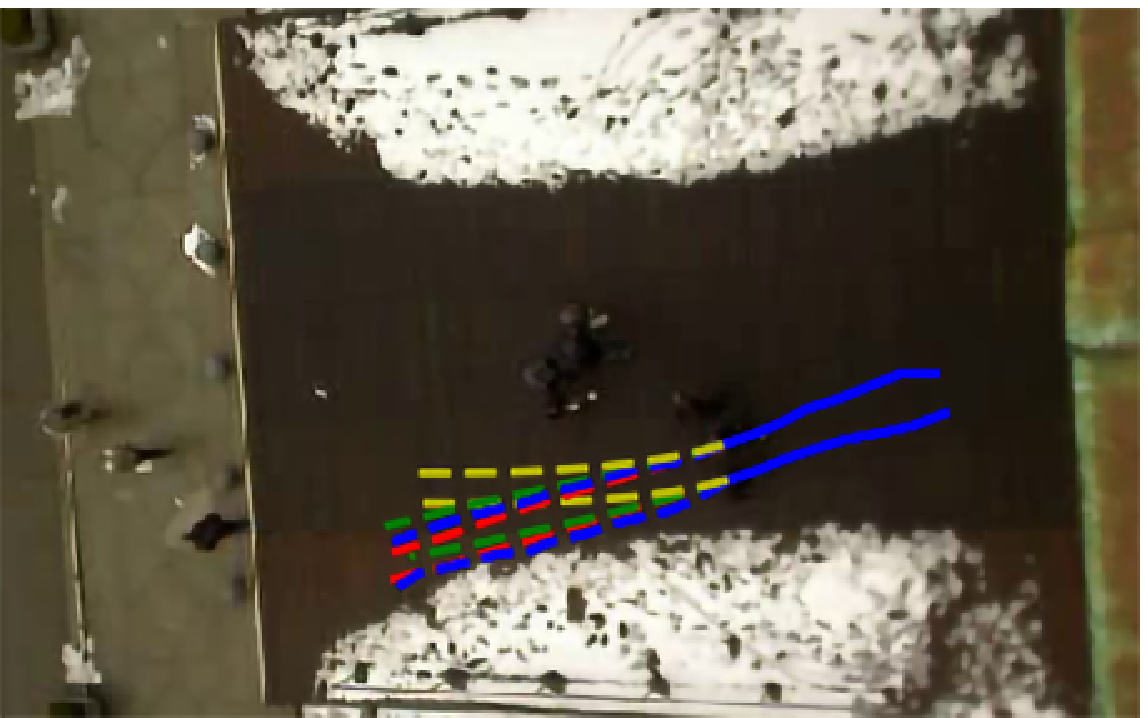}
		\end{minipage}
		\begin{minipage}[c]{0.23\linewidth}
			\centerline{ETH-hotel\_\#10480}
			\vspace{0.05cm}
			\includegraphics[scale=0.34]{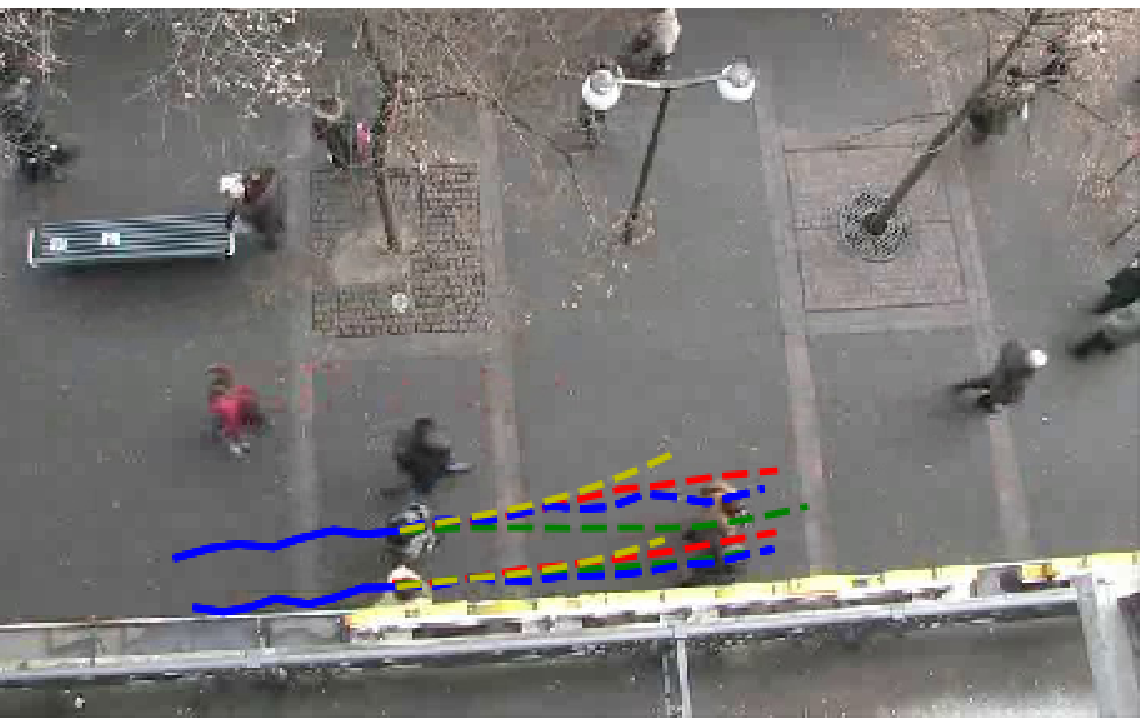}
		\end{minipage}
		\begin{minipage}[c]{0.23\linewidth}
			\centerline{UCY-zara1\_\#3700}
			\vspace{0.05cm}
			\includegraphics[scale=0.34]{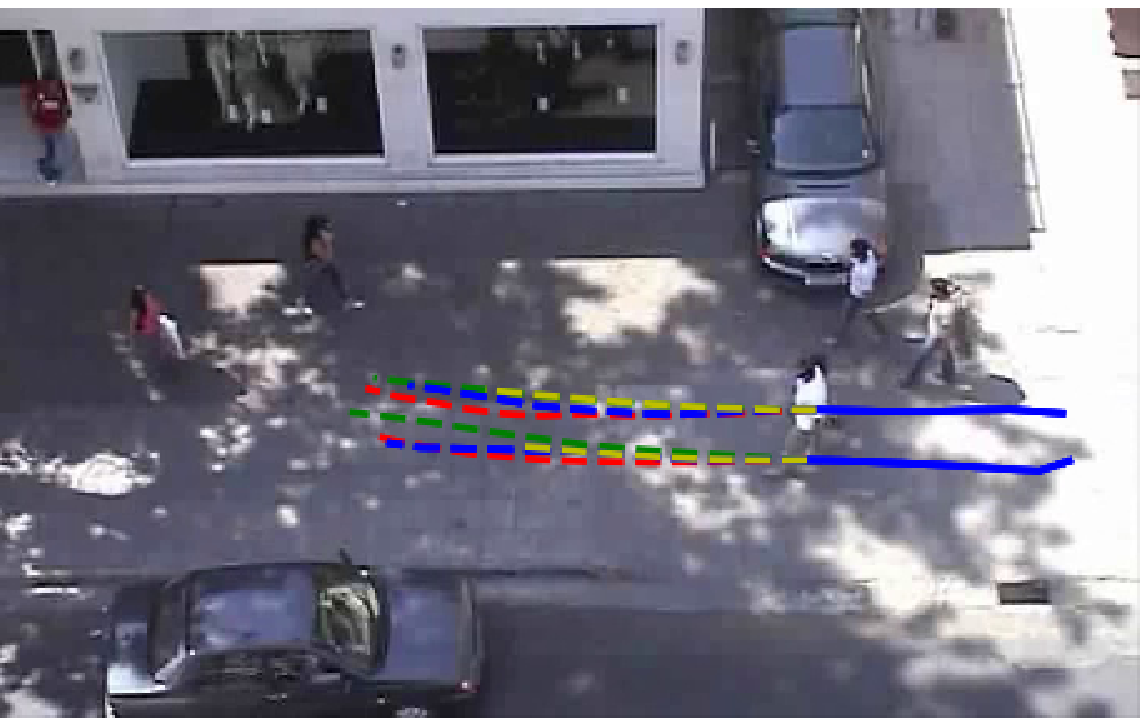}
		\end{minipage}
		\begin{minipage}[c]{0.23\linewidth}
			\centerline{UCY-univ\_\#4380}
			\vspace{0.05cm}
			\includegraphics[scale=0.34]{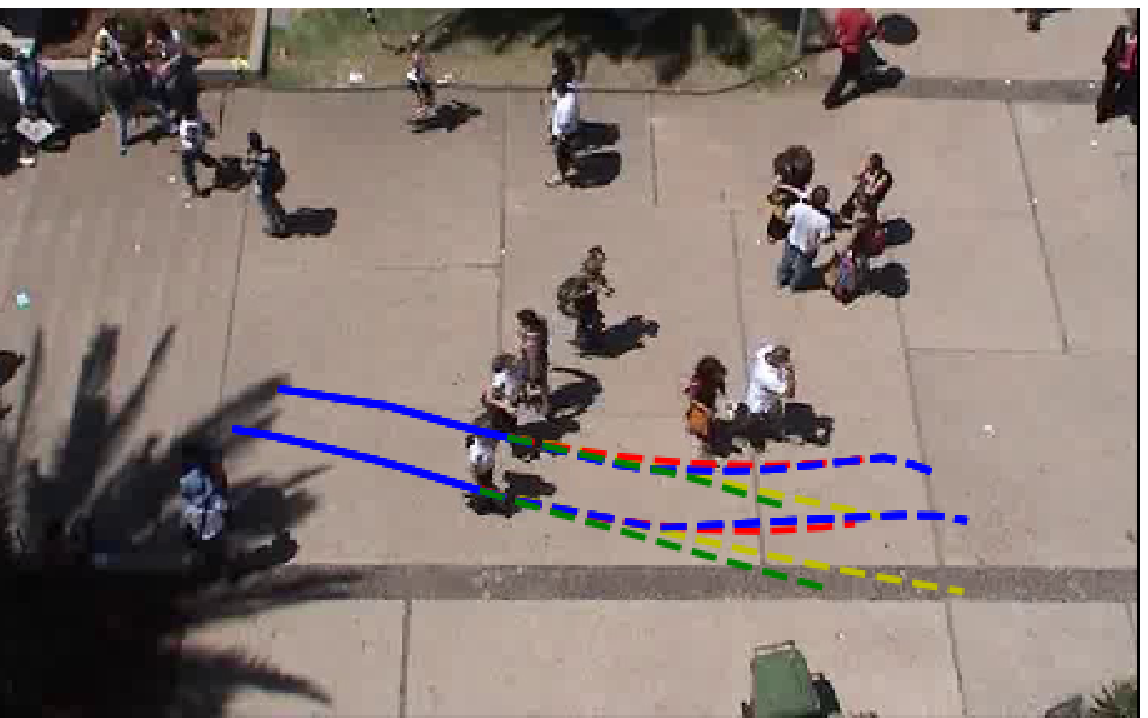}
		\end{minipage}
		\label{fig:5a}
	}
	\centering
	\subfigure[People Merging]
	{
		\centering
		\begin{minipage}[c]{0.23\linewidth}
			\centerline{ETH-univ\_\#4871}
			\vspace{0.05cm}
			\includegraphics[scale=0.34]{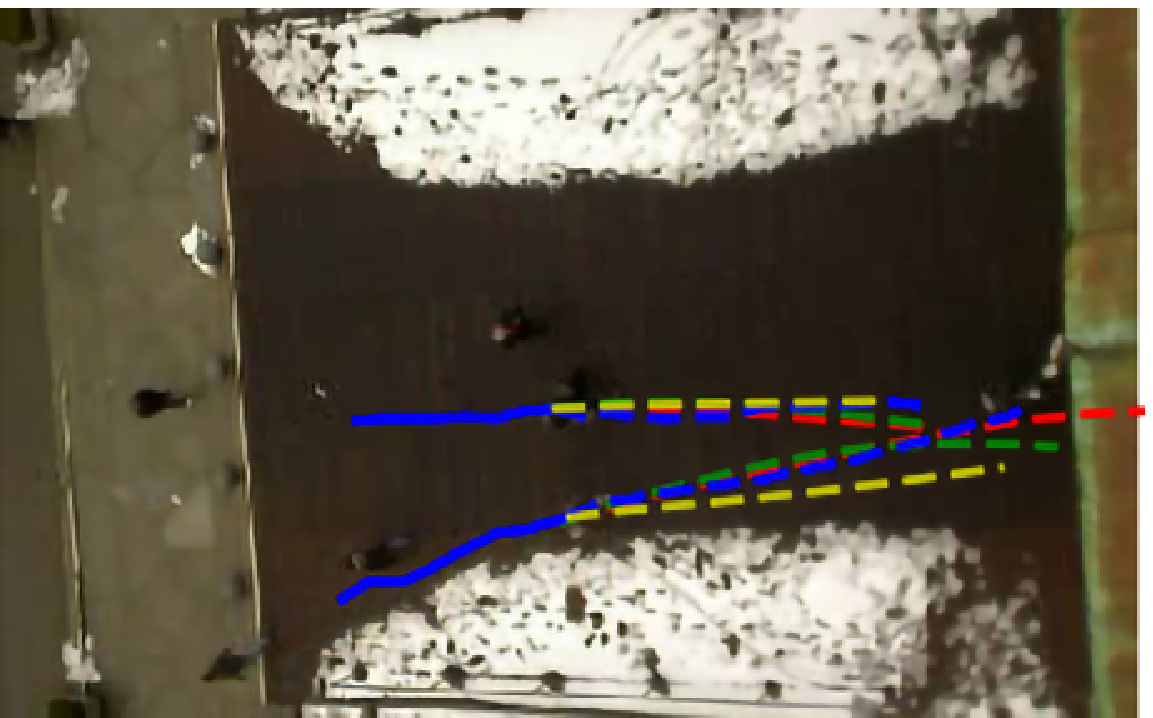}
		\end{minipage}
		\begin{minipage}[c]{0.23\linewidth}
			\centerline{UCY-zara1\_\#3460}
			\vspace{0.05cm}
			\includegraphics[scale=0.34]{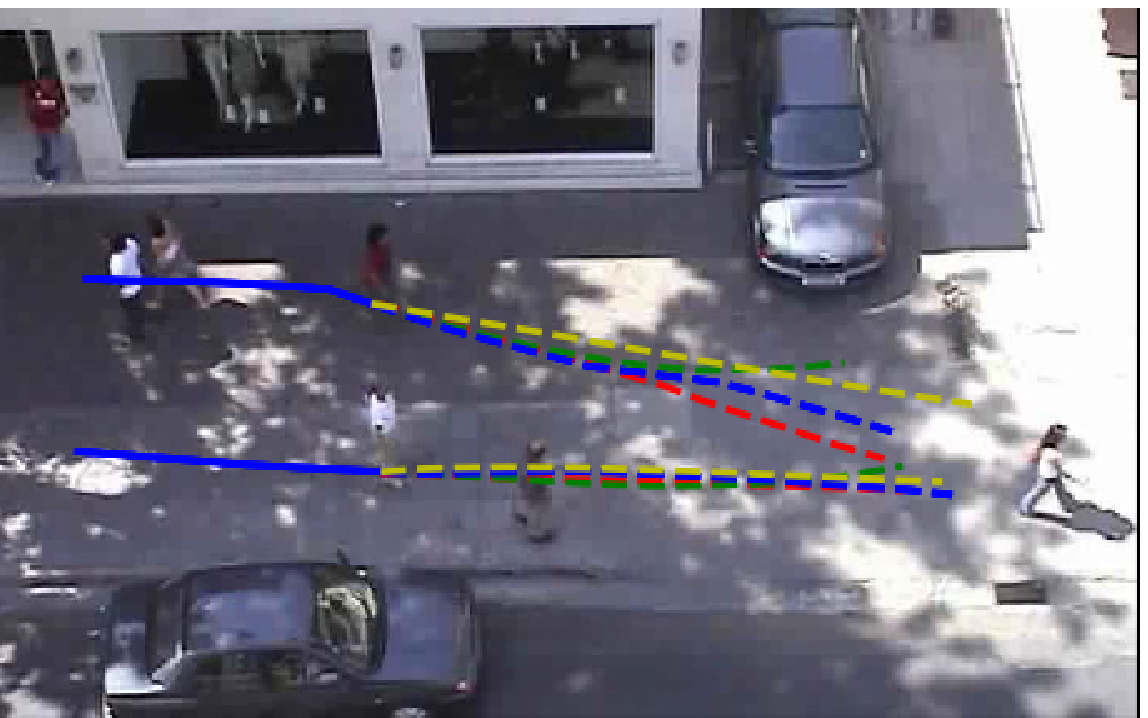}
		\end{minipage}
		\begin{minipage}[c]{0.23\linewidth}
			\centerline{UCY-zara2\_\#500}
			\vspace{0.05cm}
			\includegraphics[scale=0.34]{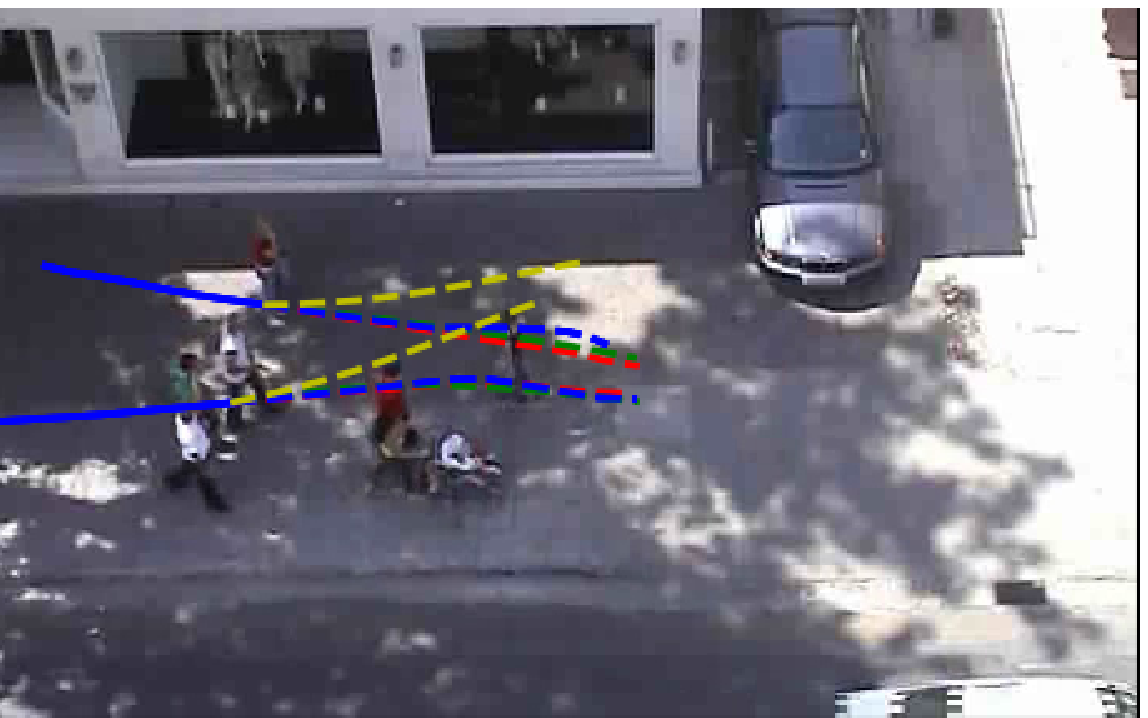}
		\end{minipage}
		\begin{minipage}[c]{0.23\linewidth}
			\centerline{UCY-zara2\_\#6010}
			\vspace{0.05cm}
			\includegraphics[scale=0.34]{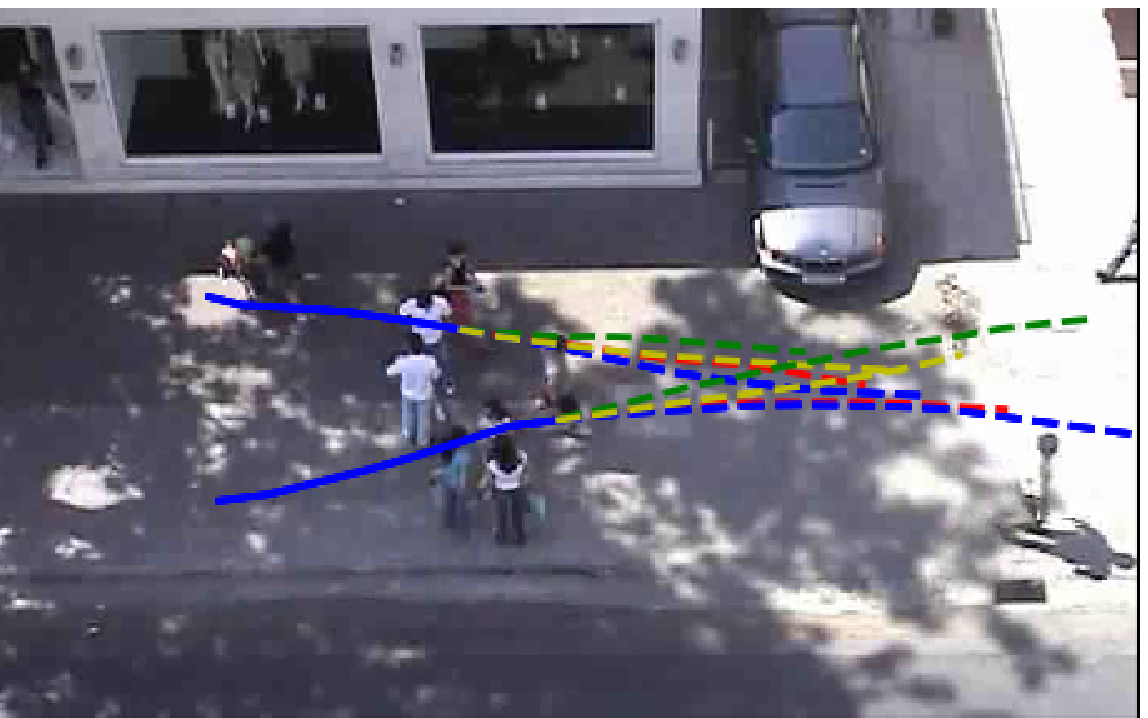}
		\end{minipage}
		\label{fig:5b}
	}
	\centering
	\subfigure[Person Following]
	{
		\begin{minipage}[c]{0.23\linewidth}
			\centerline{ETH-univ\_\#12303}
			\vspace{0.05cm}
			\includegraphics[scale=0.34]{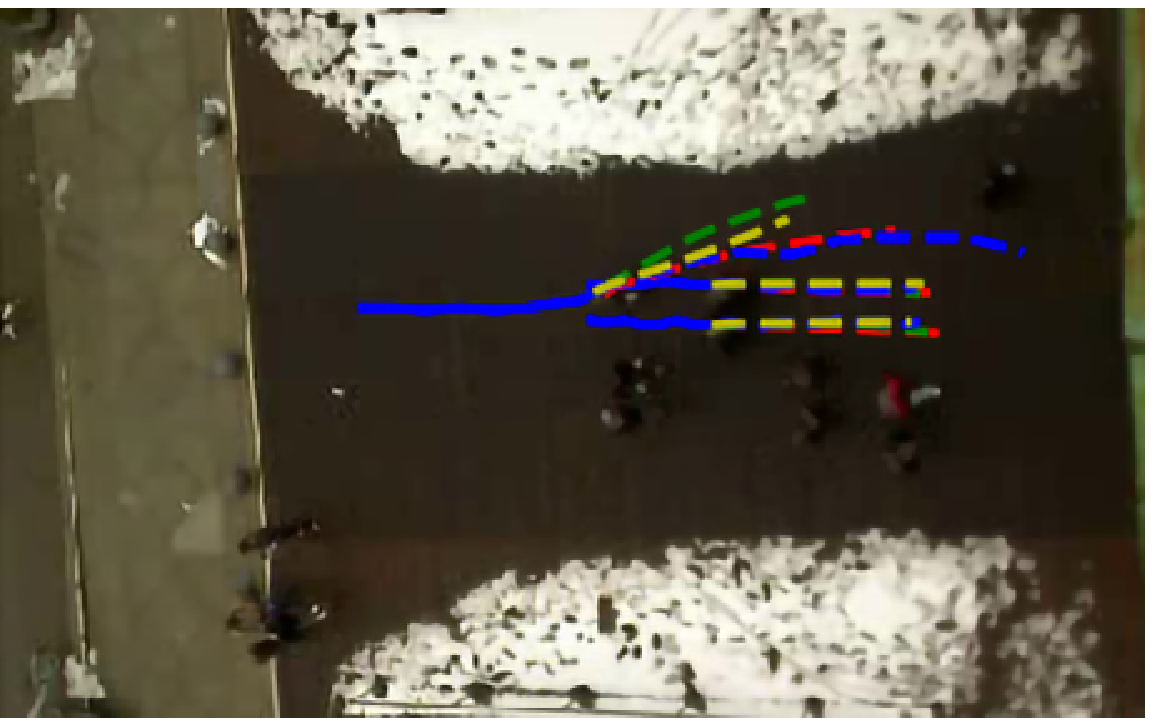}
		\end{minipage}
		\begin{minipage}[c]{0.23\linewidth}
			\centerline{UCY-zara1\_\#7770}
			\vspace{0.05cm}
			\includegraphics[scale=0.34]{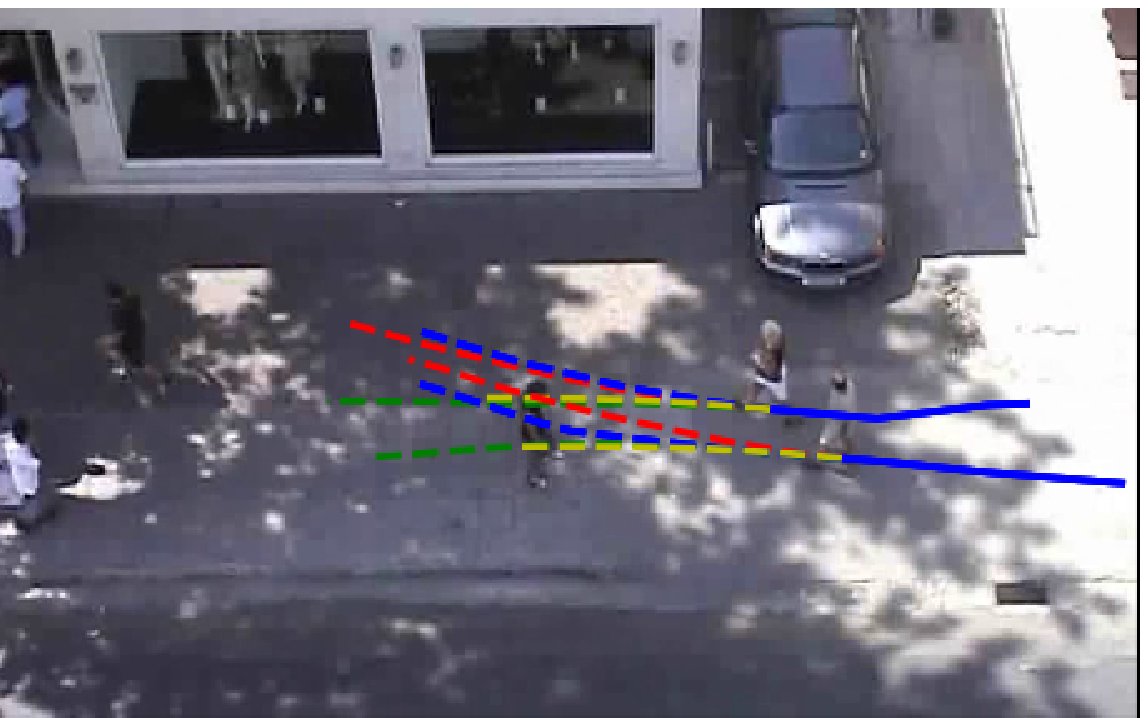}
		\end{minipage}
		\begin{minipage}[c]{0.23\linewidth}
			\centerline{UCY-zara2\_\#2510}
			\vspace{0.05cm}
			\includegraphics[scale=0.34]{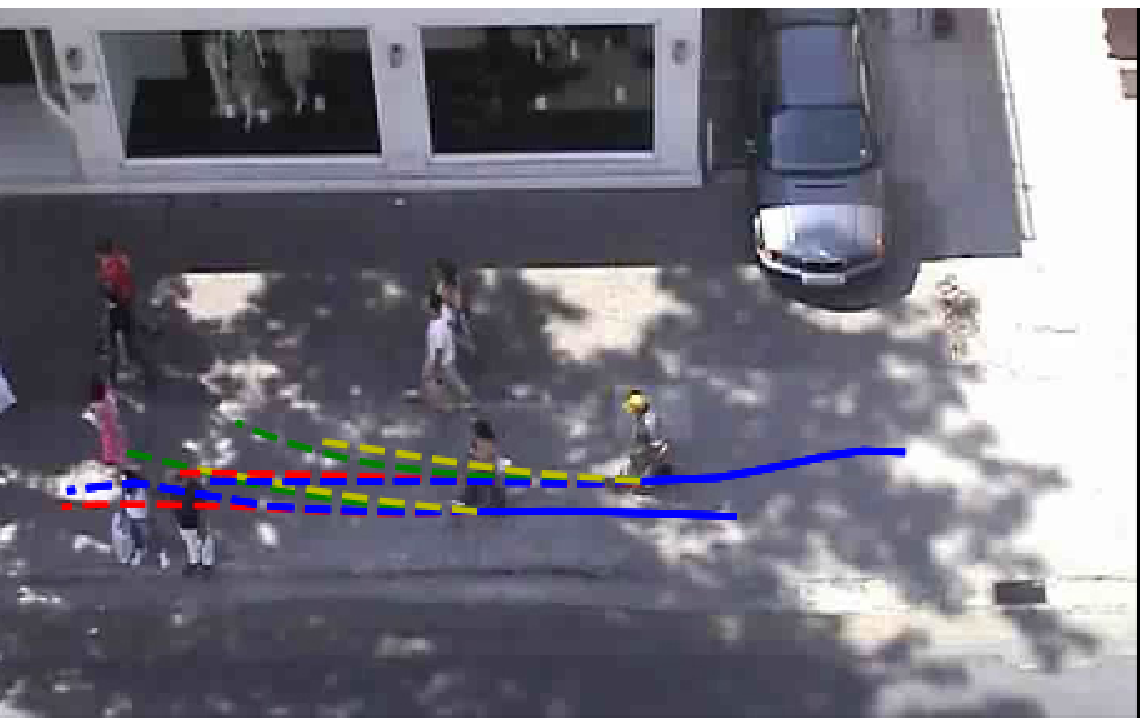}
		\end{minipage}
		\begin{minipage}[c]{0.23\linewidth}
			\centerline{UCY-zara2\_\#10320}
			\vspace{0.05cm}
			\includegraphics[scale=0.34]{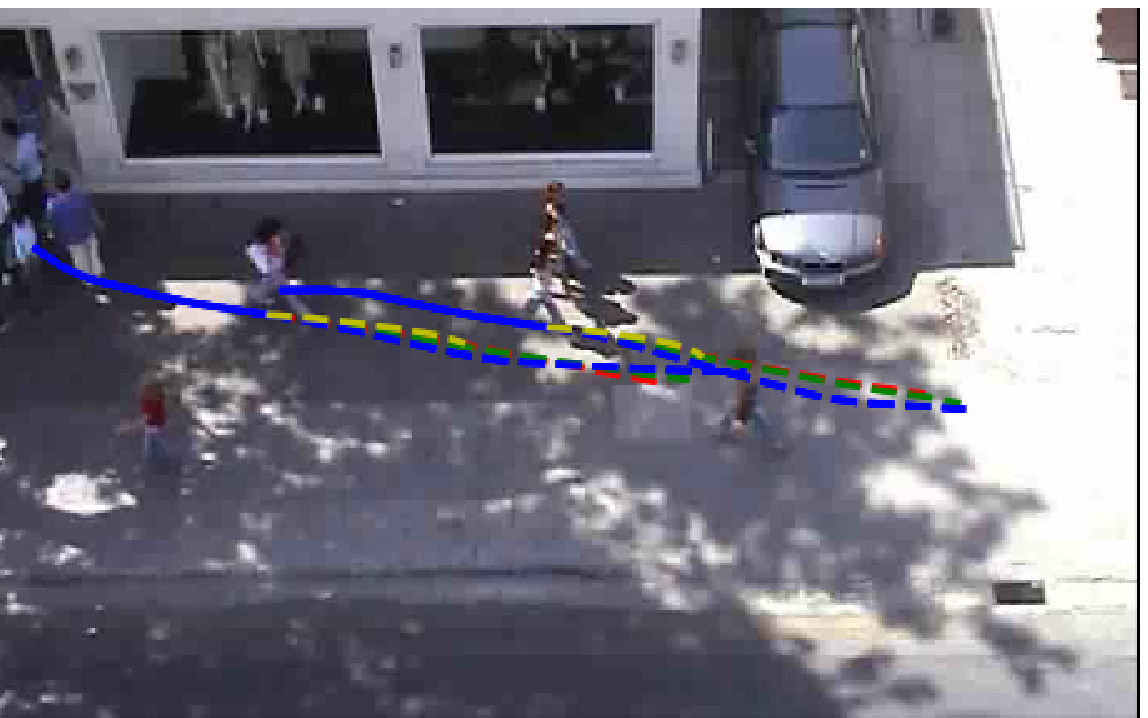}
		\end{minipage}
		\label{fig:5c}
	}
	\centering
	\subfigure[People Meeting]
	{
		\begin{minipage}[c]{0.23\linewidth}
			\centerline{ETH-univ\_\#7823}
			\vspace{0.05cm}
			\includegraphics[scale=0.34]{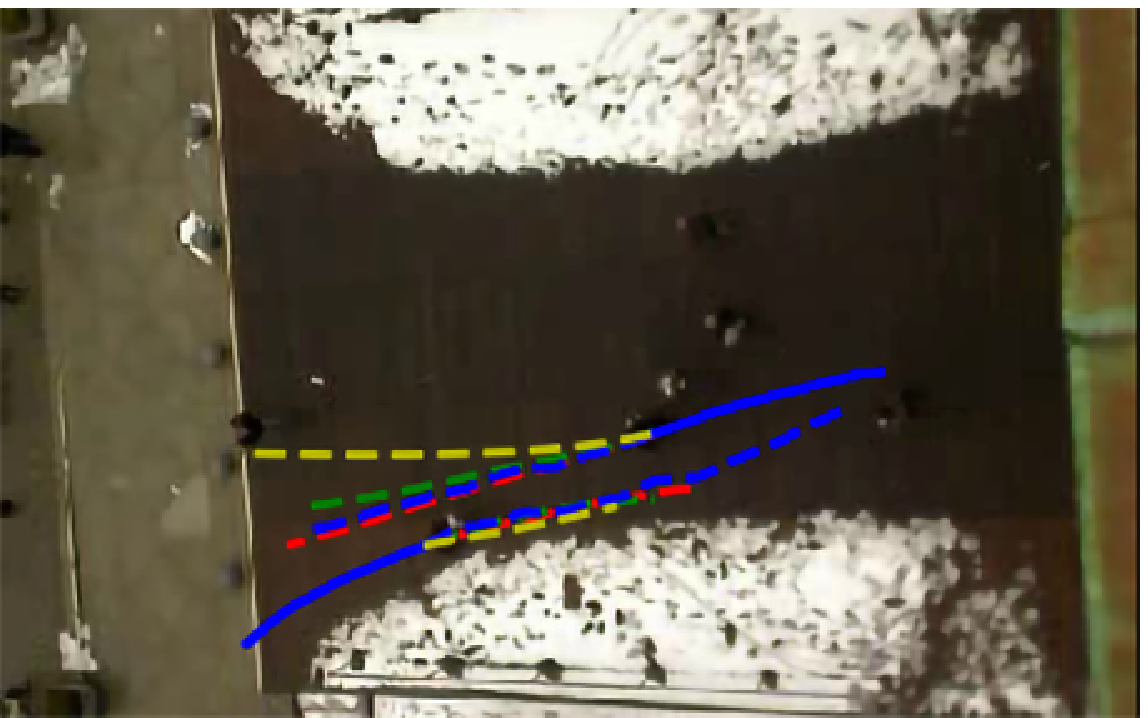}
		\end{minipage}
		\begin{minipage}[c]{0.23\linewidth}
			\centerline{UCY-zara1\_\#1030}
			\vspace{0.05cm}
			\includegraphics[scale=0.34]{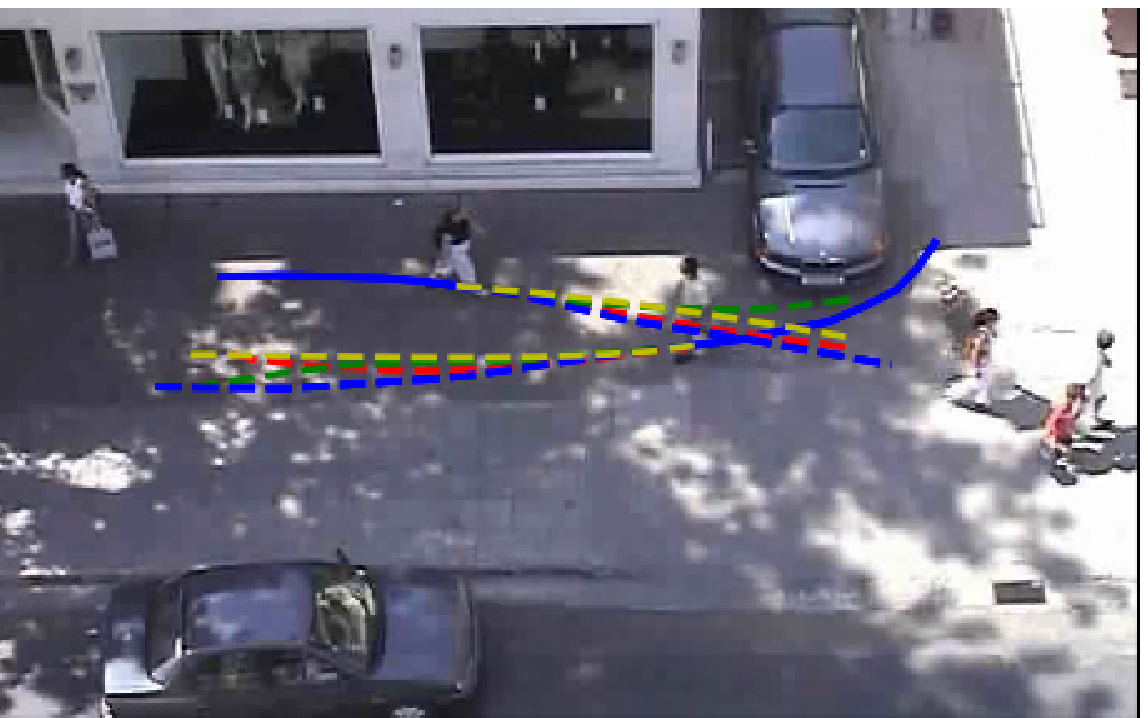}
		\end{minipage}
		\begin{minipage}[c]{0.23\linewidth}
			\centerline{UCY-zara1\_\#6890}
			\vspace{0.05cm}
			\includegraphics[scale=0.34]{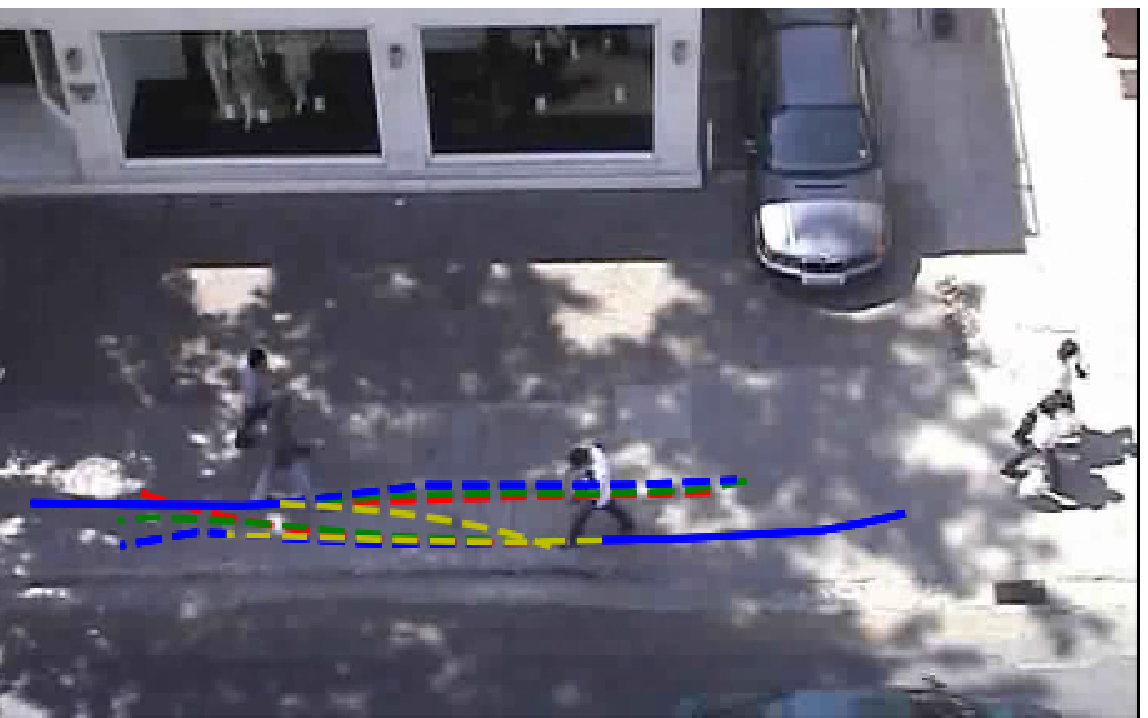}
		\end{minipage}
		\begin{minipage}[c]{0.23\linewidth}
			\centerline{UCY-univ\_\#4150}
			\vspace{0.05cm}
			\includegraphics[scale=0.34]{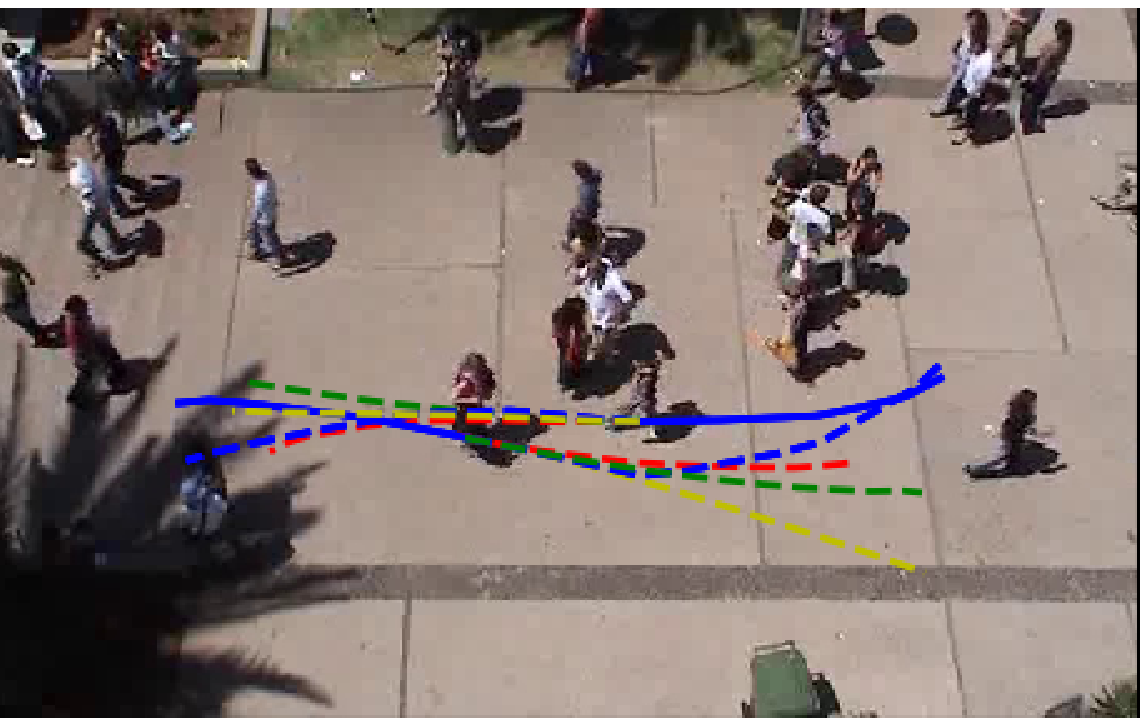}
		\end{minipage}
		\vspace{3pt}
		\label{fig:5d}
	}
	\caption{Comparisons of our method with SGAN and SR-LSTM in four common social scenarios. The examples in each row belong to the same type of social scenario, from top to bottom: parallel walking, people merging, person following, and people meeting. For each case, such as ETH-univ\_\#4277 which is the scenario of the $4277^{th}$ frame in the ETH-univ dataset, the blue solid line represents the trajectory of observed 8 key frames, the dashed line is the trajectory of future 12 key frames (blue: ground truth, yellow: SGAN, green: SR-LSTM, red: our model).}
	\label{Fig:5}
\end{figure*}

\subsubsection{Individual Interaction Scenarios}\label{subsec4.3.1}
We consider four types of social scenarios where people have to consider possible social interactions to avoid collision (see Fig. \ref{Fig:5}).

\textbf{Parallel Walking.} (Row 1) It is very common for two individuals walking in parallel in real-scenarios. For example, the behaviors of classmates going to school together and couples shopping together. There are 4 parallel walking scenarios from ETH-univ, ETH-hotel, UCY-zara01 and UCY-univ datasets shown in Fig. \ref{fig:5a}. In these scenarios, the trajectories predicted by SRA-LSTM are closest to ground truth, followed by the SR-LSTM model. For the first scenario, the SGAN model cannot successfully predict future trajectories. It is worth noting the fourth scenario, the pair of pedestrians changed their original directions of movement in future trajectory. The SGAN model and SR-LSTM model still predict the future trajectories in the original direction of movement. Our model successfully learned the change of pedestrians' movement directions and predicted the future trajectories. Unfortunately, our model does not learn the speed of pedestrian's movement that the predicted position of the final time-step is still a little bit away from ground truth. To a certain extent, our prediction is still successful.

\textbf{People Merging.} (Row 2) In hallways or on roads, it is common for people coming from different directions to merge and walk towards a common destination. People adopt various ways such as walking speed, slow down to avoid colliding while continuing towards their destinations. As the first scenario shown in Fig. \ref{fig:5b}, the lower pedestrian walk speed up to avoids collision with the upper pedestrian. The prediction of SGAN model shows that the lower person avoid collision by changing the direction of movement, which is inconsistent with ground truth. The prediction of SR-LSTM model shows that the lower pedestrian approaches to the upper pedestrian and then adjusts the direction of movement to move away, which is also inconsistent with ground truth. We predict the speed behavior and successfully predict the future trajectory, which closely matches with the ground truth trajectory. In the other three scenarios, the predicted trajectories by SRA-LSTM also closely match the ground truth unlike the deviation we see in SGAN and SR-LSTM.

\textbf{Person Following.} (Row 3) People tend to follow the people in front when walking toward a common destination. But if the person in front walks slowly, we have to bypass them to reach the destination. For the first scenario shown in Fig. \ref{fig:5c}, the three pedestrians have the common destination, but the walking speed of the pair of pedestrians in front is slower than the pedestrian behind. In this situation, the pedestrian behind chose to pass over them from the left side of the pair. For the pedestrian behind, the predicted trajectories of these three models have all detoured, but only our prediction is close to the ground truth. Although the predicted trajectory does not reach the specified position, the predicted movement direction is correct. In the other three scenarios, the trajectories predicted by our model are also the closest to the ground truth. For the two scenarios in the middle, the predictions of SGAN and SR-LSTM show that there are various deviations between the predicted trajectories and ground truth.

\textbf{People Meeting.} (Row 4) On the way to the destination, there will be other people coming from the opposite direction. For the face-to-face meeting, one has to adjust the direction of movement temporarily to avoid collision. In other cases, it is not necessary. As the first scenario shown in Fig. \ref{fig:5d}, without avoidance, the two pedestrians would pass by. The upper pedestrian's trajectories predicted by SGAN and SR-LSTM changed the direction of movement to avoid collision, which deviated from the ground truth. The trajectory predicted by our model of this person is closest to the ground truth. But the predicted trajectories for the lower pedestrian are all shorter than ground truth. For several other scenarios, our model can always predict the trajectory closest to ground truth.

\subsubsection{Group Interaction Scenarios}\label{subsec4.3.2}
When there are many pedestrians, they unconsciously form a group, especially acquaintances. We compare our model with the SR-LSTM model in two common group scenarios. The scenarios include group walking and group avoiding.

\setcounter{subfigure}{-1}
\begin{figure}[!t]
	\centering
	\subfigure
	{
		\centering
		\includegraphics[scale=0.3]{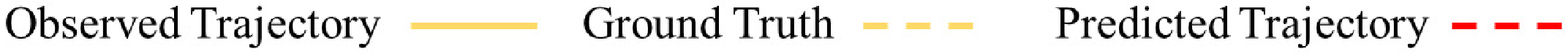}
	}
	\\
	\subfigure[]
	{
		\begin{minipage}[c]{0.45\linewidth}
			\centering
			\centerline{ETH-univ\_\#10395}
			\vspace{0.05cm}
			\includegraphics[scale=0.175]{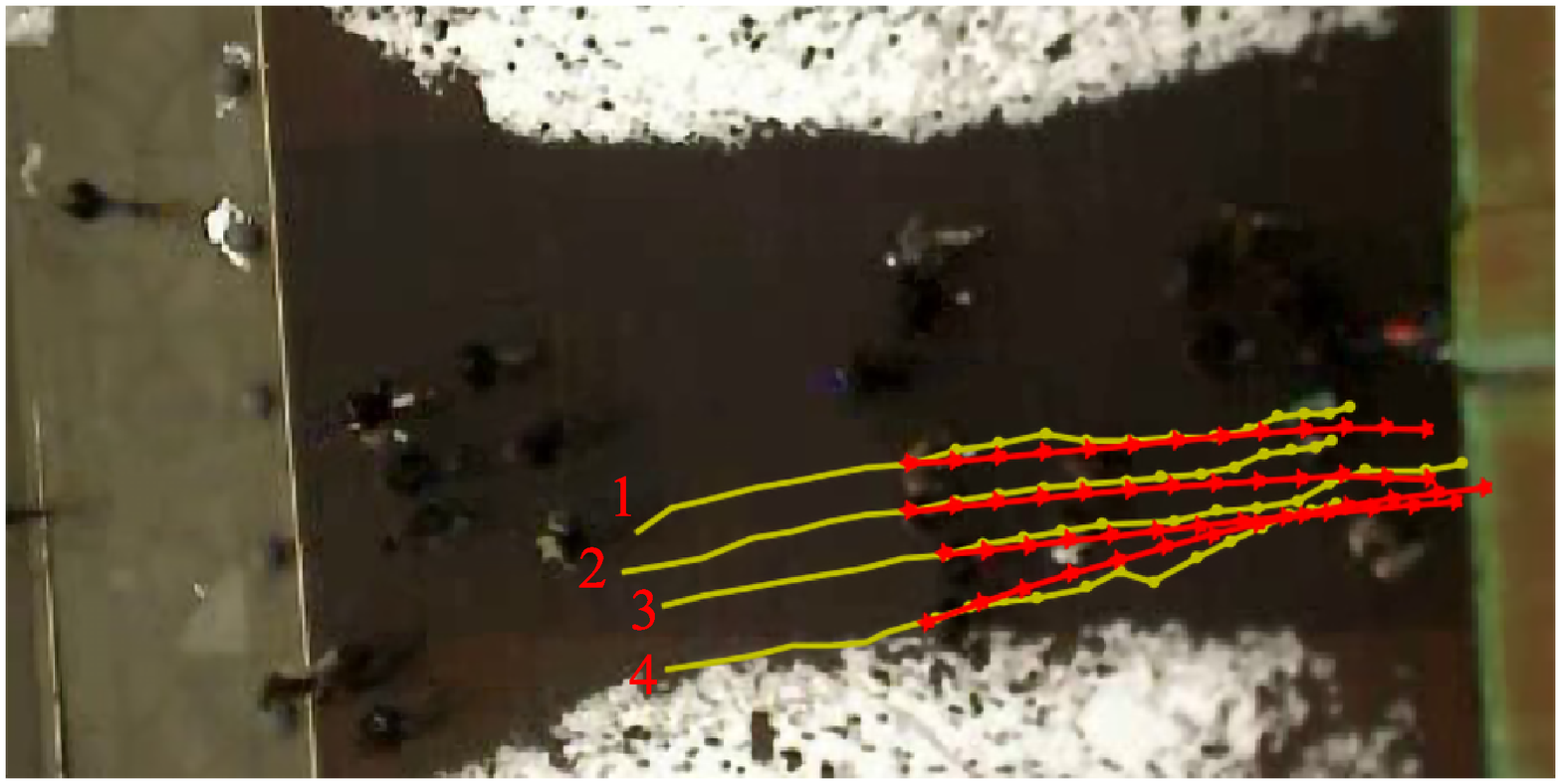} \\
			\vspace{3pt}
			\centerline{UCY-zara1\_\#7500}
			\vspace{0.05cm}
			\includegraphics[scale=0.175]{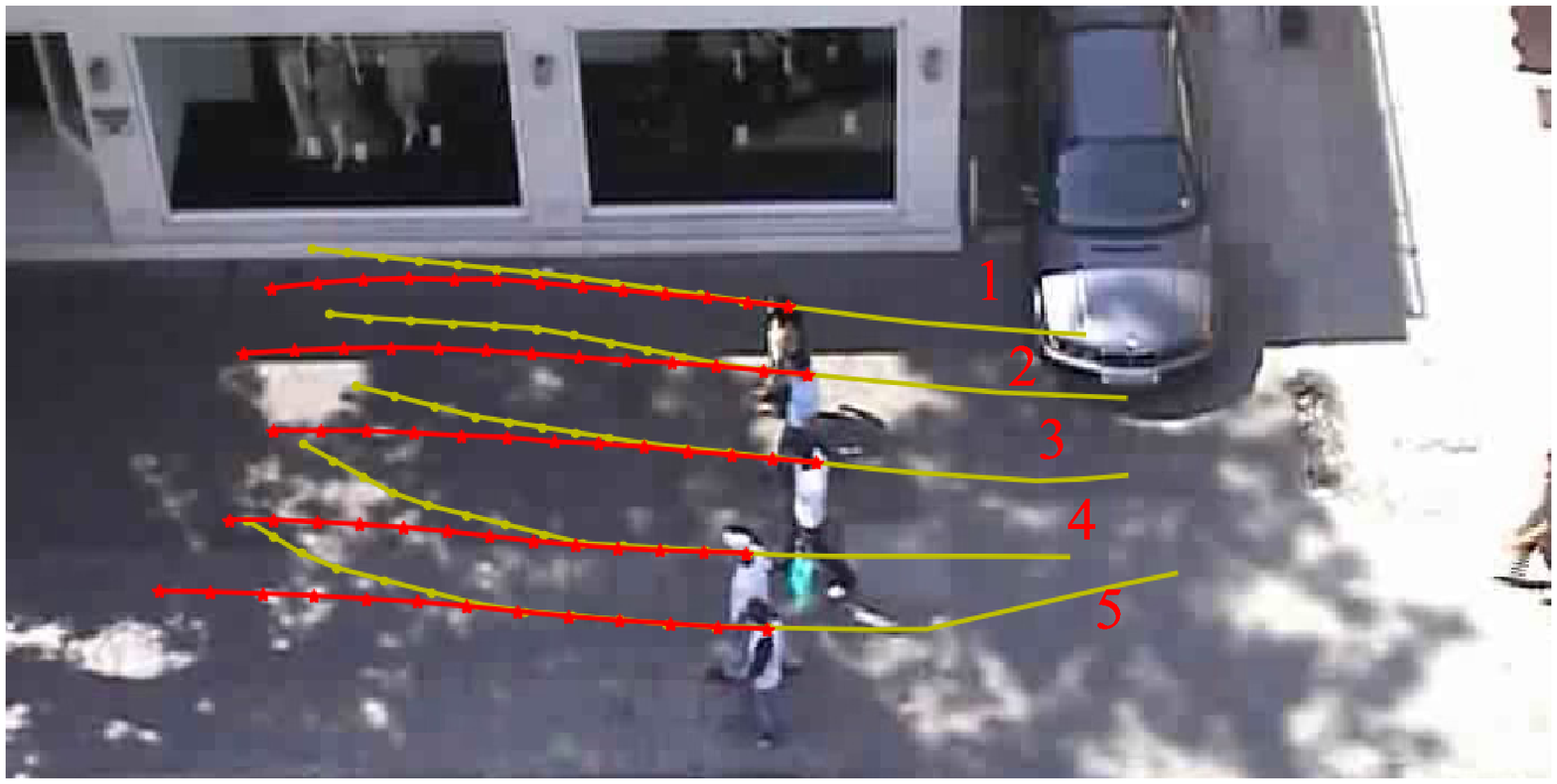} \\
			\vspace{3pt}
		\end{minipage}
		\label{fig:6a}
	}
	\subfigure[]
	{
		\begin{minipage}[c]{0.45\linewidth}
			\centering
			\centerline{ETH-univ\_\#10395}
			\vspace{0.05cm}
			\includegraphics[scale=0.175]{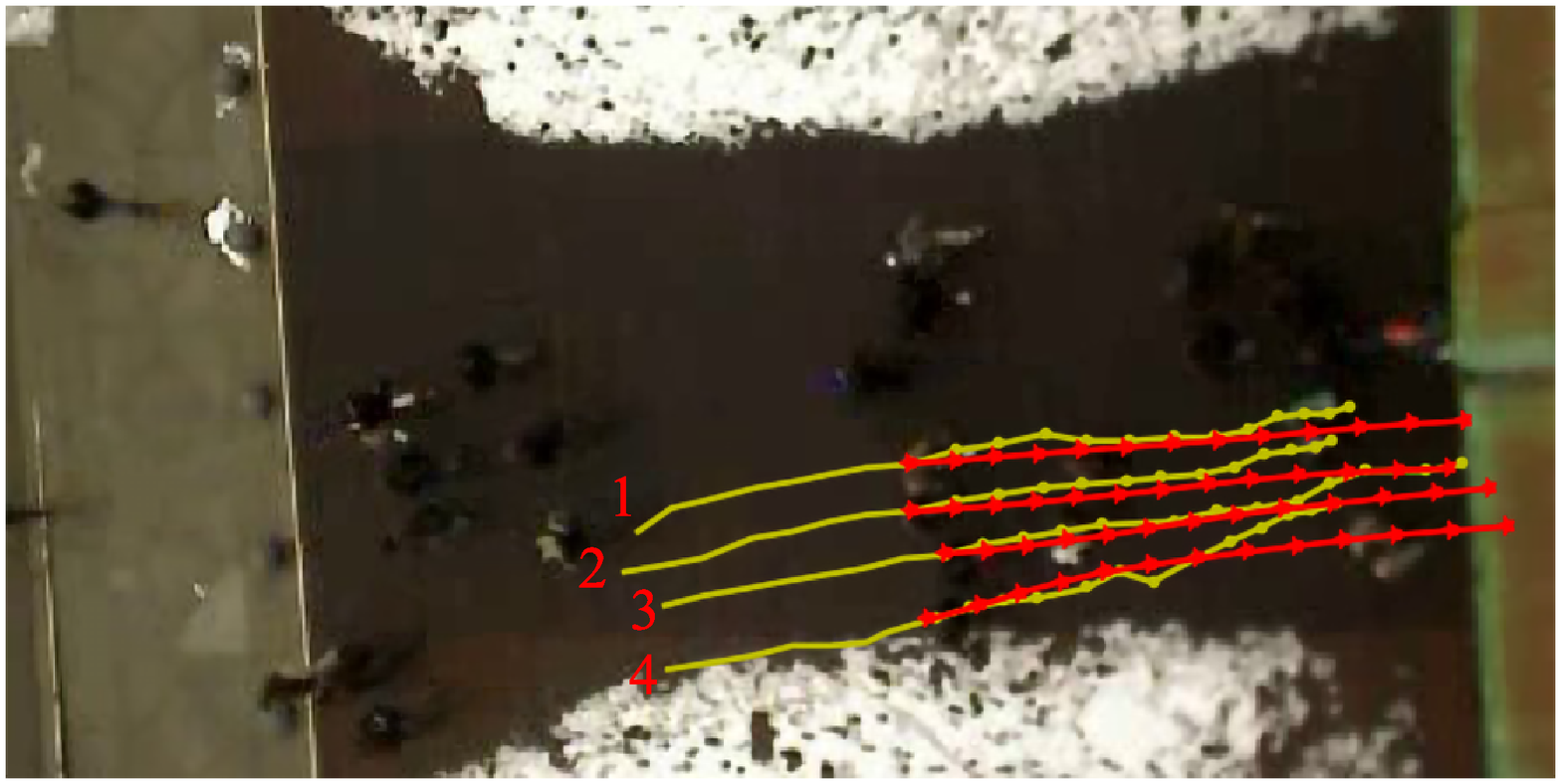} \\
			\vspace{3pt}
			\centerline{UCY-zara1\_\#7500}
			\vspace{0.05cm}
			\includegraphics[scale=0.175]{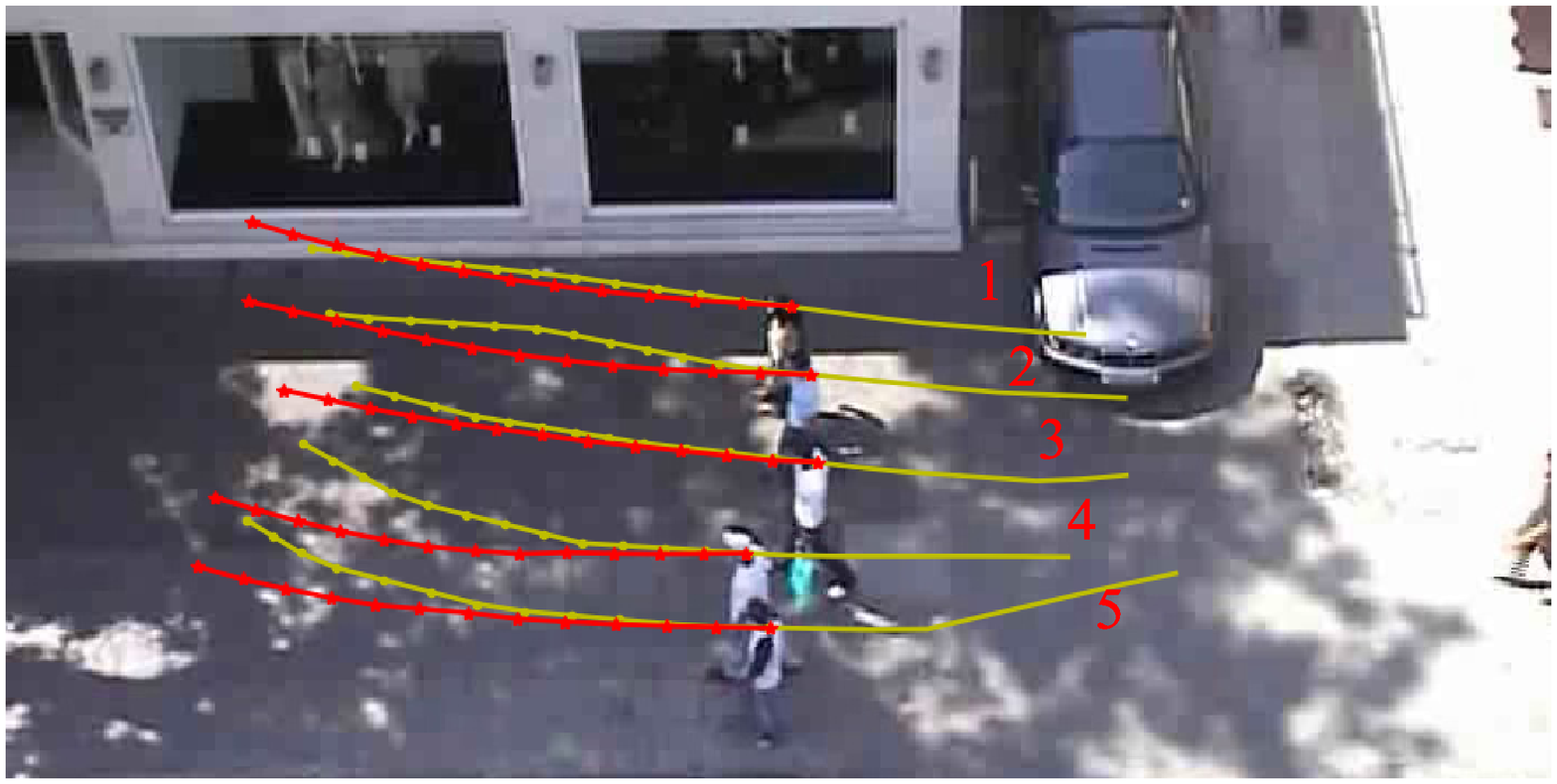} \\
			\vspace{3pt}
		\end{minipage}
		\label{fig:6b}
	}
	\caption{Scenarios of group walking parallel case. (a) shows the predictions of SR-LSTM model. (b) shows the predictions of SRA-LSTM model.}
	\label{Fig:6}
\end{figure}

\setcounter{subfigure}{-1}
\begin{figure}[!t]
	\centering
	\subfigure
	{
		\centering
		\includegraphics[scale=0.3]{fig7.eps}
	}
	\\
	\subfigure[]
	{
		\begin{minipage}[c]{0.45\linewidth}
			\centering
			\centerline{UCY-zara1\_\#1660}
			\vspace{0.05cm}
			\includegraphics[scale=0.175]{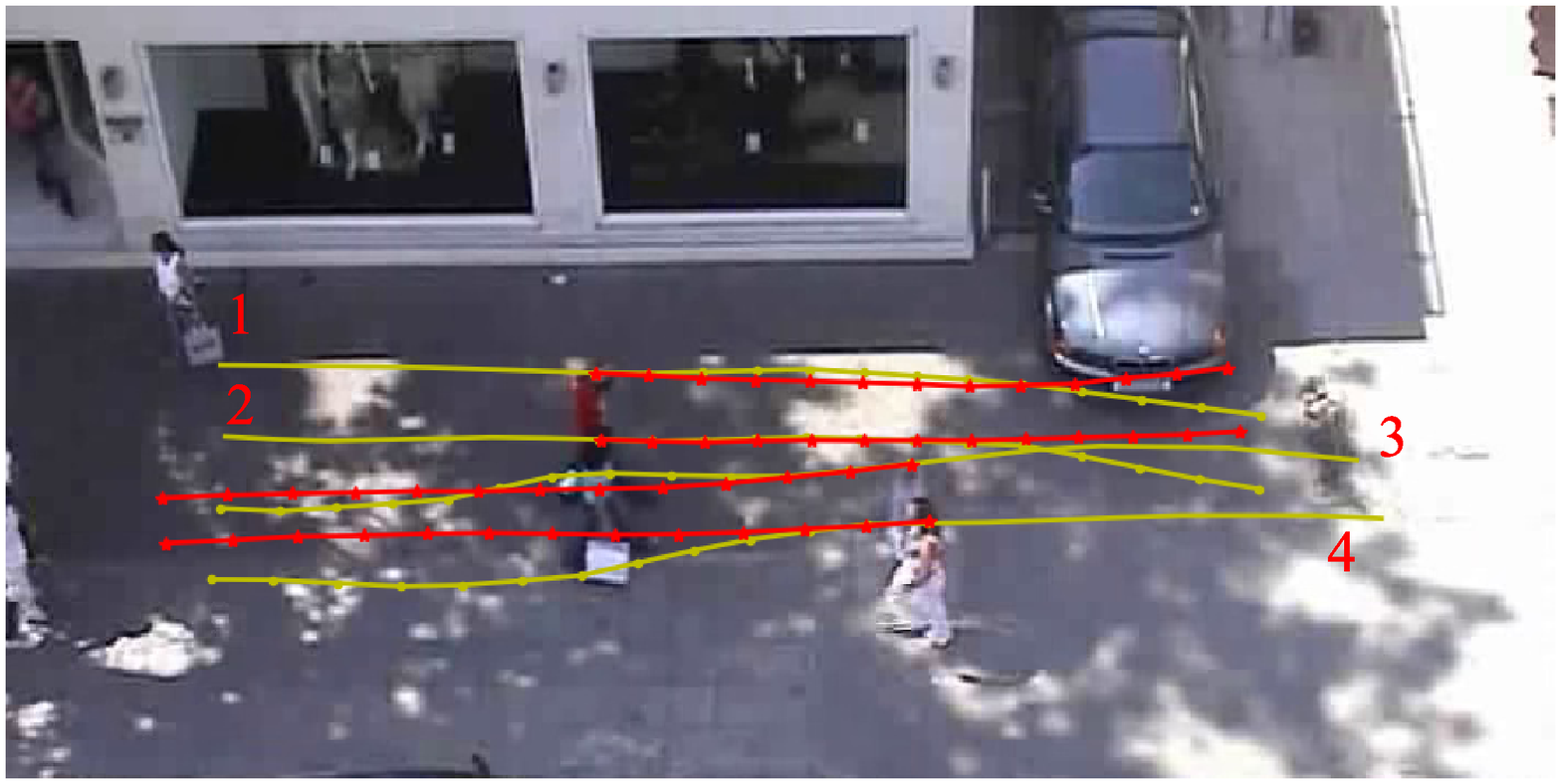} \\
			\vspace{3pt}
			\centerline{UCY-zara2\_\#1320}
			\vspace{0.05cm}
			\includegraphics[scale=0.175]{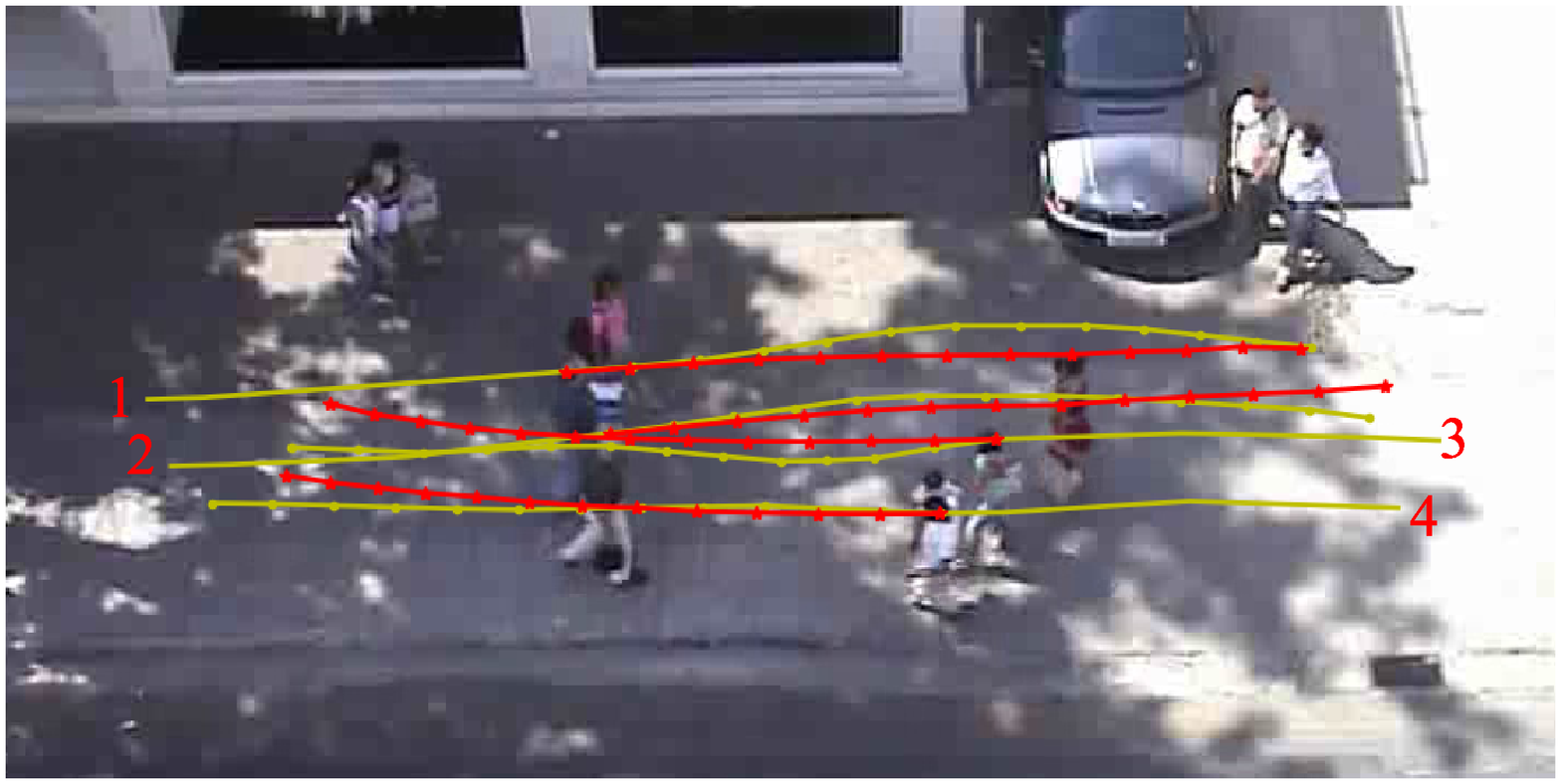} \\
			\vspace{3pt}
		\end{minipage}
		\label{fig:7a}
	}
	\subfigure[]
	{
		\begin{minipage}[c]{0.45\linewidth}
			\centering
			\centerline{UCY-zara1\_\#1660}
			\vspace{0.05cm}
			\includegraphics[scale=0.175]{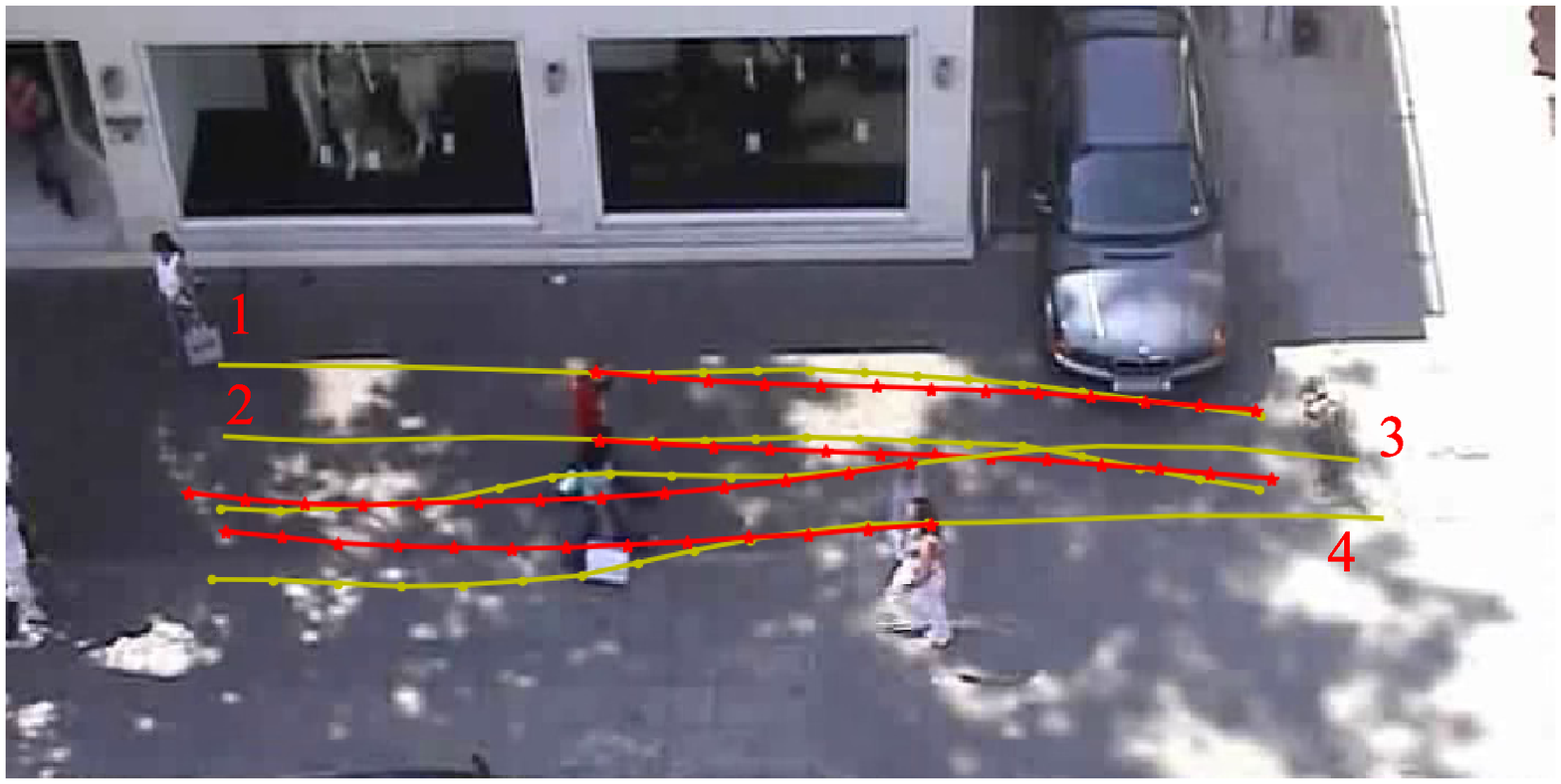} \\
			\vspace{3pt}
			\centerline{UCY-zara2\_\#1320}
			\vspace{0.05cm}
			\includegraphics[scale=0.175]{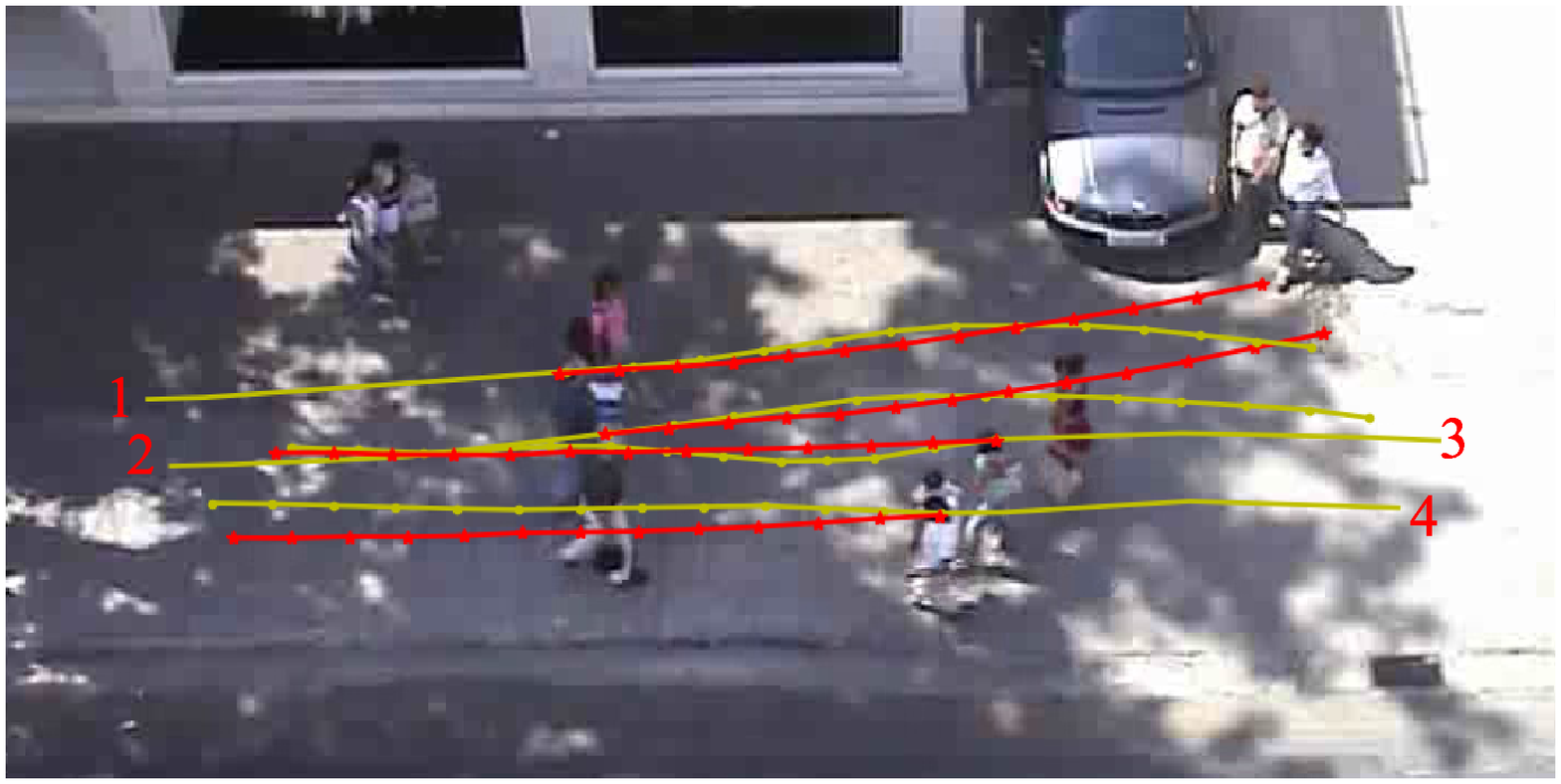} \\
			\vspace{3pt}
		\end{minipage}
		\label{fig:7b}
	}
	\caption{Scenarios of group avoiding case. (a)shows the prediction of SR-LSTM model. (b)shows the prediction of SRA-LSTM model.}
	\label{Fig:7}
\end{figure}

\textbf{Group walking.} The common group walking behaviors such as several classmates going to school together or several friends shopping together. The group walking scenarios from ETH-univ and ETH-zara01 datasets are shown in Fig. \ref{Fig:6}. Column 1 shows the prediction of SR-LSTM model, and column 2 shows the prediction of our model. For the upper scenario, For the scenario shown in row 1, the trajectories predicted by SR-LSTM show that pedestrians 3 and 4 are gradually getting closer together, which means that the two collide. Although there is a deviation between the trajectory predicted by our model and the ground truth, there is no collision. For the scenario in row 2, there are five pedestrians walking together. There are various deviations between the trajectories predicted by SR-LSTM with the ground truth. The same goes for our predicted trajectories. Although our predicted trajectories also have various deviations, they are closer to ground truth. It also shows that our predicted pedestrian's walking speed is faster than in ground truth.

\textbf{Group avoiding.} There are two scenarios of group avoiding cases shown in Fig. \ref{Fig:7}. In this scenario, two pairs of pedestrians met from opposite directions. For the scenario in row 1, pedestrians 1 and 2's trajectories predicted by our model are closer to ground truth than SR-LSTM. Pedestrian 3 changed the direction of movement to avoid collision with pedestrian 2, and pedestrian 4 as a partner of pedestrian 3 also changed. The pedestrian 3's trajectory predicted by our model is closer to the ground truth. The trajectories of pedestrian 4 predicted by the two models are quite different from the ground truth. For the scenario in row 2, the pair of pedestrians 1 and 2 change the direction of their movement to avoid collision with the other pair, and then return to the original direction. Our prediction only changed the direction of pedestrians 1 and 2, but did not return to the original direction, which is the reason for the deviation of the prediction trajectory. However, the predicted trajectory of pedestrian 3 matches with ground truth very well. The trajectory of pedestrian 4 predicted by SR-LSTM is better than our prediction.

\subsubsection{Failure Cases Scenarios}\label{subsec4.3.3}
\begin{figure}[!t]
	\centering
	\subfigure[UCY-zara2\_\#1320]{
		\centering
		\includegraphics[width=3.8cm]{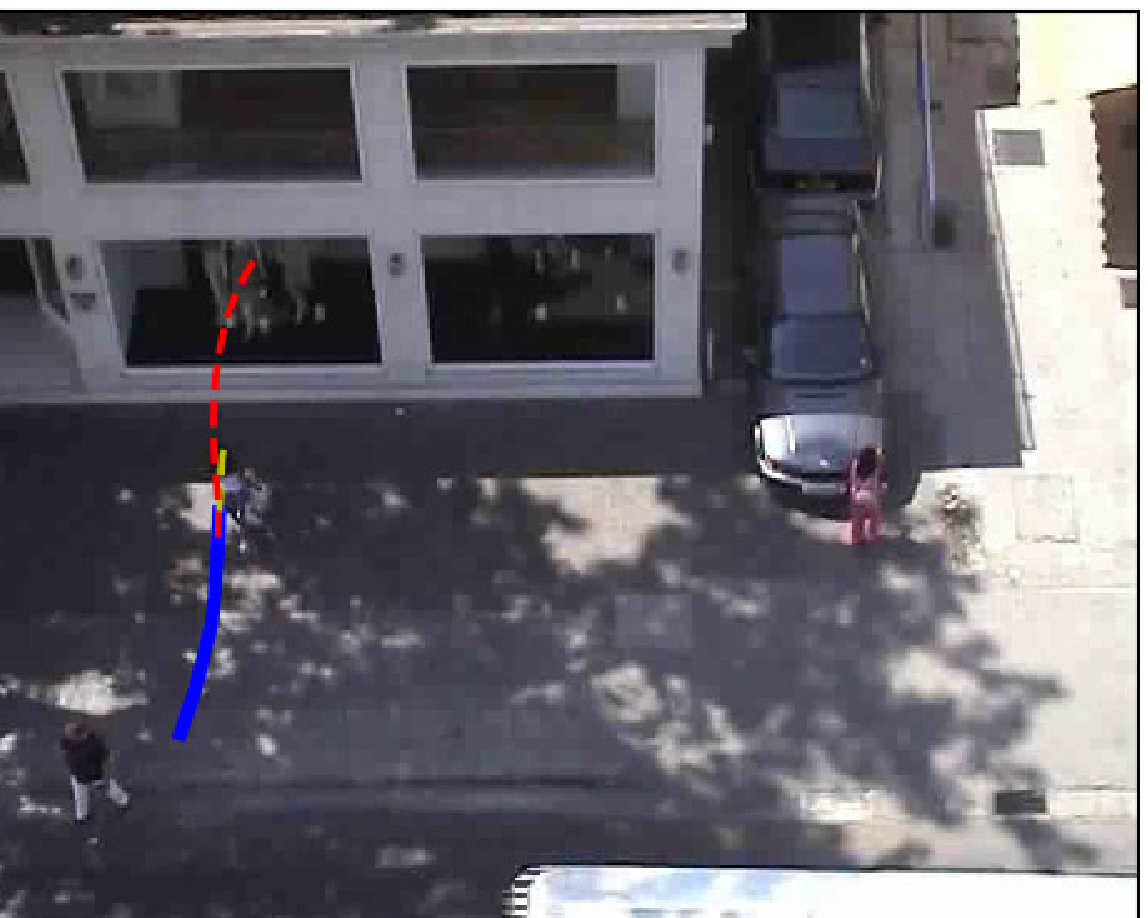}
		\label{Fig:8a}}
	\subfigure[UCY-zara2\_\#2730]{
		\centering
		\includegraphics[width=3.8cm]{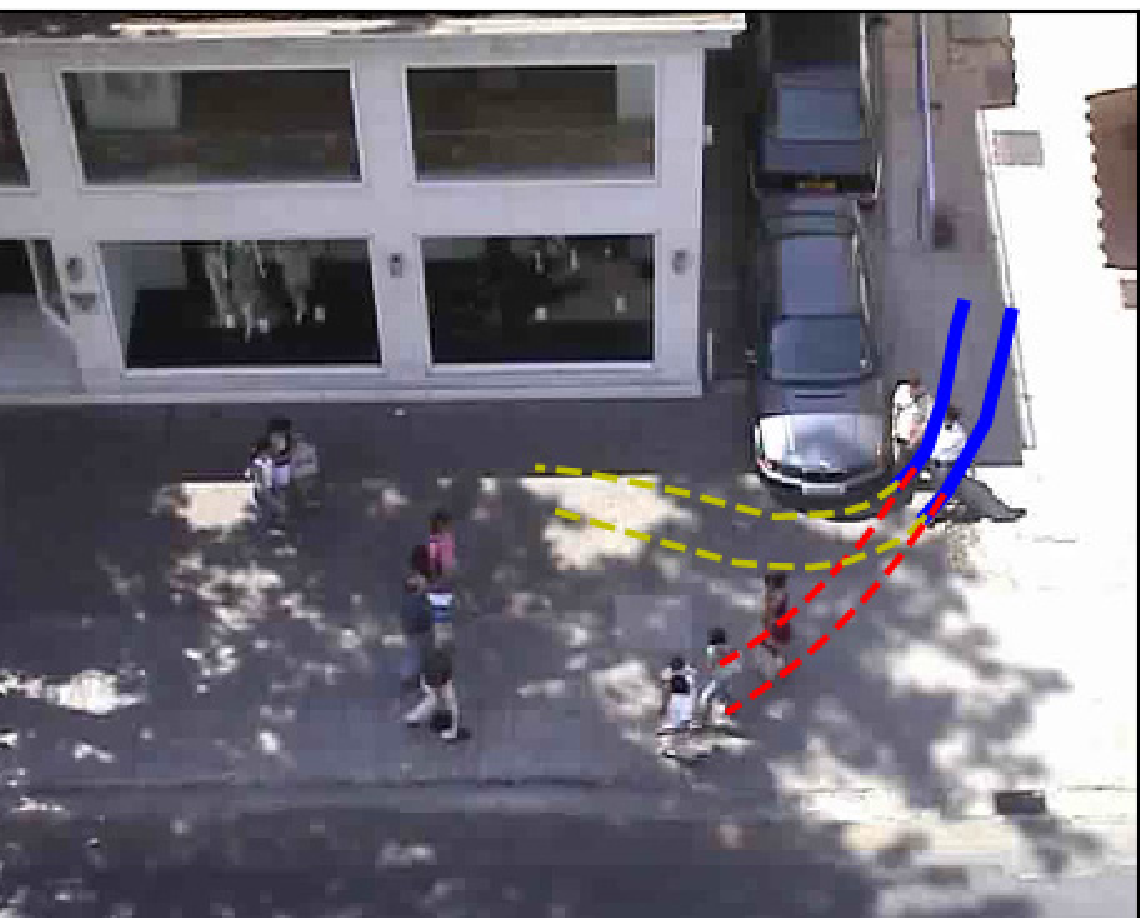}
		\label{Fig:8b}}
	\centering
	\caption{Scenarios of failure case. The blue solid line represents the observed trajectory, the yellow dashed line represents groundtruth, and the red dashed line represents the predicted trajectory.}
	\label{Fig:8}
\end{figure}

The above two subsections show the successful cases of the model in individual interaction scenarios and group interaction scenarios. However, our model cannot successfully predict the future trajectory of pedestrians in some special scenarios. Two failure case scenarios are shown in Fig. \ref{Fig:8}. The target pedestrian in Fig. \ref{Fig:8a} walks towards the wall and chooses to stop by the wall. Our model cannot understand this motivation from the pedestrian's observed trajectory, and thus incorrectly predicts the future trajectory. The reason for the failure case shown in Fig. \ref{Fig:8b} is also that the model cannot understand the pedestrian's motivation from the observed trajectory, which leads to the failure of the prediction. If the model can understand the scene structure information, the above failure cases may not exist. However, the thorough solution to the problem lies in accurately understanding the motion motivation of pedestrians, which is also a difficulty in the trajectory prediction research. And it is also the goal of our future work.
\section{Conclusions and Future Work}\label{sec5}
In this paper, we propose a novel interactive recurrent structure SRA-LSTM for jointly predicting the future trajectories of pedestrians in the crowd. We also introduce a social relationship attention, which is acquired from social relationship representation between pedestrians and their latent motion patterns. The social relationship attention is adopt to aggregate movement information from neighbor pedestrians to model social interactions. The advantage of this method is that it simultaneously takes into account the impact of potential movement interactions and social relationships on motion decision. Quantitative evaluations on public datasets prove that our model is superior to state-of-art models. The qualitative results of some individual interaction scenarios and group interaction scenarios demonstrate the effectiveness of our method in social interaction modeling.

Although our SRA-LSTM model outperforms other methods in social interaction modeling, the scene layout also impacts the movement of pedestrians. In the future, we will focus on modeling human-scene interactions. We intend to capture scene layout features from a semantic map of the scene, and then combine it with human motion information to model the potential interactions between humans and the scene. We expect to improve the prediction performance by coupling human-scene interaction with our SRA-LSTM prediction model. Besides, how to effectively perceive the intentions of pedestrians' movement is also a difficult point that we need to solve. This paper focus on pedestrian trajectory prediction for surveillance video. In the future, we also hope to study pedestrian trajectory prediction based on the first perspective in order to apply to robot navigation system.

\begin{acknowledgements}
This work is supported by the National Natural Science Foundation of China (61972128), the Fundamental Research Funds for the Central Universities of China (Grant No.PA2019GDPK0071).
\end{acknowledgements}

\section*{Conflict of interest}
The authors declare that they have no conflict of interest.

\end{document}